\documentclass[10pt,twocolumn,letterpaper]{article}

\usepackage[pagenumbers]{cvpr} %

\usepackage[accsupp]{axessibility}  %
\usepackage{pifont}
\newcommand{\cmark}{\ding{51}} %
\newcommand{\xmark}{\ding{55}} %
\usepackage[table]{xcolor}
\usepackage{tikz}

\usepackage{microtype}

\usepackage{pgfplots}
\pgfplotsset{compat=1.18}
\usepgfplotslibrary{groupplots}
\definecolor{cvprblue}{rgb}{0.21,0.49,0.74}
\usepackage[pagebackref,breaklinks,colorlinks,allcolors=cvprblue]{hyperref}

\usepackage{makecell}

\def\paperID{41404} %
\def\confName{CVPR}
\def\confYear{2026}

\title{Scene-VLM: Multimodal Video Scene Segmentation via Vision-Language Models}

\author{
Nimrod Berman$^{1*\dagger}$ \quad Adam Botach$^{2*\S\ddagger}$ \quad Emanuel Ben-Baruch$^{2}$ \quad Shunit Haviv Hakimi$^{2}$ \\
Asaf Gendler$^{2}$ \quad Ilan Naiman$^{1\dagger}$ \quad Erez Yosef$^{3\dagger}$ \quad Igor Kviatkovsky$^{2}$ \\[0.5em]
$^{1}$Ben-Gurion University \quad $^{2}$Amazon Prime Video \quad $^{3}$Tel-Aviv University \\[0.3em]
{\tt\footnotesize \{bermann,naimani\}@post.bgu.ac.il \quad \{kabotach,emanbb,havivs,gendlasa,kviat\}@amazon.com \quad erez.yo@gmail.com}
}

\newcommand{\methodname}{\emph{Scene-VLM}}

\begin{document}
\maketitle

\renewcommand{\thefootnote}{}
\footnotetext{$^{*}$Equal contribution.}
\footnotetext{$^{\dagger}$Work done during an Amazon internship.} \footnotetext{$^{\S}$Internship mentor.}
\footnotetext{$^{\ddagger}$Corresponding author: \href{mailto:kabotach@amazon.com}{\textcolor{black}{\texttt{kabotach@amazon.com}}}.}
\renewcommand{\thefootnote}{\arabic{footnote}}

\begin{abstract}
Segmenting long-form videos into semantically coherent scenes is a fundamental task in large-scale video understanding. Existing encoder-based methods are limited by visual-centric biases, classify each shot in isolation without leveraging sequential dependencies, and lack both narrative understanding and explainability.
In this paper, we present Scene-VLM, the first fine-tuned vision-language model (VLM) framework for video scene segmentation.
Scene-VLM jointly processes visual and textual cues including frames, transcriptions, and optional metadata to enable multimodal reasoning across consecutive shots.
The model generates predictions sequentially with causal dependencies among shots and introduces a context–focus window mechanism to ensure sufficient temporal context for each shot-level decision. In addition, we propose a scheme to extract confidence scores from the token-level logits of the VLM, enabling controllable precision–recall trade-offs that were previously limited to encoder-based methods. 
Furthermore, we demonstrate that our model can be aligned to generate coherent natural-language rationales for its boundary decisions through minimal targeted supervision.
Our approach achieves state-of-the-art performance on standard scene segmentation benchmarks. On MovieNet, for example, \methodname~ yields significant improvements of +6 AP and +13.7 F1 over the previous leading method.

\end{abstract}

\vspace{-6mm}
\section{Introduction}
\begin{figure}
    \centering
    \includegraphics[width=0.9\linewidth]{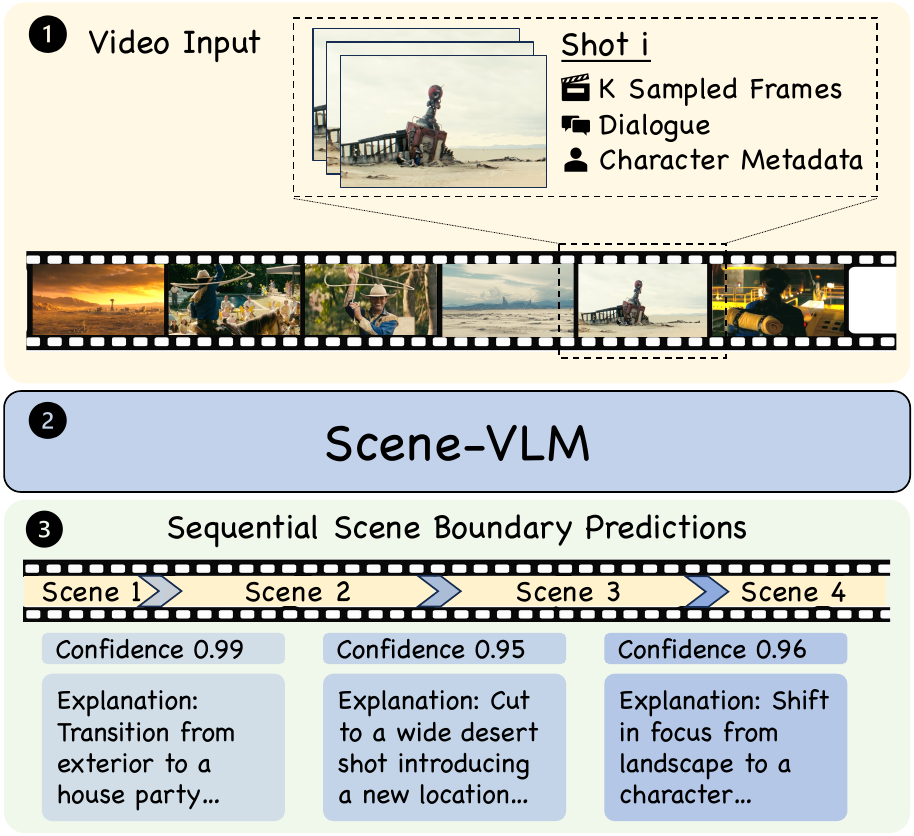}
    \caption{\textbf{Video scene segmentation with Scene-VLM.} 
    We present Scene-VLM, the first vision-language model (VLM) framework fine-tuned for video scene segmentation. Scene-VLM jointly processes visual frames, dialogue, and metadata from consecutive shots to sequentially predict scene boundaries with associated confidence scores, and can be aligned to produce coherent post-hoc explanations for its decisions.}
    \label{fig:what_is_scene_seg}
    \vspace{-6mm}
\end{figure}

\vspace{-1mm}
Video scene segmentation, the task of identifying coherent narrative boundaries within long-form video content, is fundamental to video understanding~\citep{baraldi2015deep, rao2020local, wu2022scene, mun2022boundary, islam2023efficient, sadoughi2023mega, huang2020movienet}. Accurate scene boundary detection is crucial for organizing, searching, and understanding video content at scale, enabling applications such as automated structured summarization, semantic retrieval and contextual advertising. Formally, a \emph{scene} is a consecutive sequence of shots sharing semantic coherence in location, time, characters, or narrative theme, where each \emph{shot} is a continuous sequence of frames captured in a single, uninterrupted camera take (see~\cref{fig:what_is_scene_seg}). Despite decades of research, video scene segmentation remains challenging as it requires understanding narrative semantics beyond visual cues, distinguishing meaningful story transitions from superficial visual changes.

\begin{figure*}[t]
    \centering
    \includegraphics[width=0.9\linewidth]{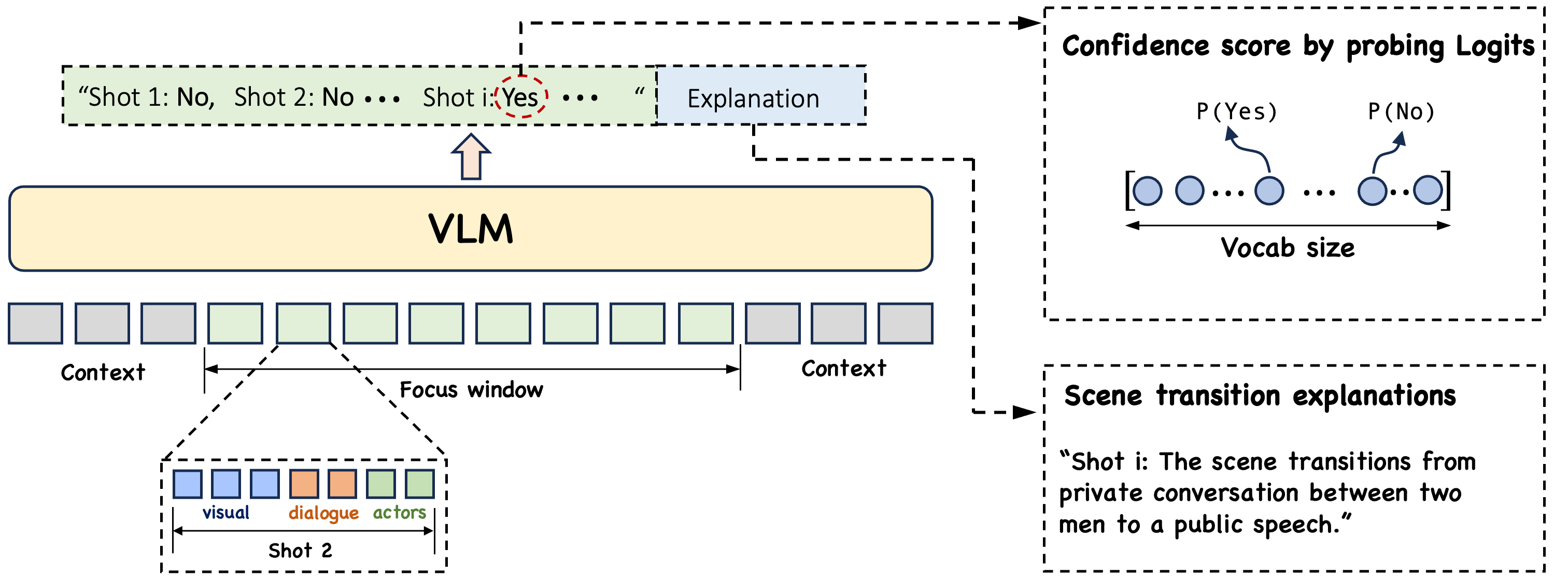}
    \caption{\textbf{Proposed approach.} The VLM receives a sequence of $N$ multimodal shot representations as input. Each shot representation consists of visual frames, dialogue, and optional metadata. 
\methodname~ processes the shots within a focus window (green) using information from a larger context window (gray), and outputs scene boundary predictions for the shots in focus in the format ``Shot~$i$: Yes/No.'' 
For each shot, we compute a confidence score by probing the softmax logits of the ``Yes'' and ``No'' tokens and normalize by 
$P(\texttt{Yes})/(P(\texttt{Yes}) + P(\texttt{No}))$. 
The model can also be aligned to generate coherent post-hoc explanations for its scene-boundary decisions.}
    \label{fig:method_flow}
    \vspace{-5mm}
\end{figure*}

Recent state-of-the-art methods have made substantial progress through cross-modal fusion and efficient temporal modeling. BaSSL~\citep{mun2022boundary} employs boundary-aware self-supervised pretraining with pseudo-boundaries to learn transition cues. In contrast, TranS4mer~\citep{islam2023efficient} combines self-attention with state-space layers to efficiently process long-range dependencies, and MEGA~\citep{sadoughi2023mega} aligns video, audio, and text via cross-modal attention. Despite these advances, existing methods share several fundamental limitations. First, they all exhibit visual-centric biases, ignoring or underutilizing non-visual signals such as dialogue and character presence despite their importance for narrative understanding~\citep{huang2020movienet}. Second, they all follow a mutual point-wise prediction paradigm, classifying each shot independently within a local temporal window without leveraging causal dependencies across consecutive decisions. Finally, as encoder-based methods, they all offer no insight into \emph{why} a boundary was predicted beyond a confidence score, limiting their usability in human-in-the-loop settings.

Recent advances in vision-language models (VLMs)~\cite{qin2024qwen2_5vl, llama4, glm4_5v, claude4, gpt4, gemini_2_5} offer a compelling path to overcome these limitations. VLMs unify visual perception and natural language understanding in a single framework, processing visuals alongside text and generating textual responses that enable deeper cross-modal reasoning and explainable predictions. Popular model families such as Qwen-VL~\cite{qin2024qwen2_5vl}, LLaMA~\citep{llama4}, and GLM~\citep{glm4_5v} demonstrate strong multimodal capabilities, making them viable backbones for structured video understanding tasks.

In this work, we introduce~\methodname, a fine-tuned VLM framework for video scene segmentation. To the best of our knowledge, this represents the first application of VLMs to this task, marking a paradigm shift from traditional encoder-based approaches to multimodal narrative reasoning. As illustrated in~\cref{fig:method_flow}, \methodname{} addresses prior limitations through several key design choices. 

First, we introduce a \textbf{structured multimodal shot representation} fusing visual frames, speech, and optional metadata, providing the model with narrative context unavailable to visual-centric methods. 
Second, we replace the common point-wise prediction paradigm with \textbf{sequential processing}, where predictions for multiple shots are generated sequentially, making each boundary decision causally inform subsequent ones. 
Third, to mitigate edge effects from limited context at sequence boundaries, we employ a \textbf{context–focus window design}: a broader context window provides temporal padding around a central focus window where predictions are emitted, ensuring every shot has adequate past and future evidence. 
Fourth, since VLMs do not natively produce confidence scores, we propose a \textbf{confidence prediction scheme} that reliably derives confidence scores from token-level logits of the model's outputs. This enables flexible operating points across precision–recall trade-offs, a capability typically reserved for encoder-based classifiers.
Finally, we demonstrate that \methodname{} can be effectively aligned to generate \textbf{coherent natural-language rationales} for its boundary predictions through fine-tuning on a small set of annotated explanations. 

To conclude, we summarize our contributions as follows:
\begin{itemize}
    \item We introduce~\methodname, a \emph{fine-tuned VLM framework for video scene segmentation}, featuring a structured multimodal shot representation, sequential predictions with causal dependencies, and a context–focus window design for comprehensive temporal reasoning.
    \item We propose a novel \emph{scene boundary confidence prediction scheme} that derives scores from VLM token-level logits, enabling controllable precision–recall trade-offs typically reserved for encoder-based methods.
    \item We achieve \emph{state-of-the-art results} on MovieNet (+6 AP, +13.7 F1 over the previous leading method), strong zero-shot performance on BBC Planet Earth, and demonstrate adaptability to 
    video chaptering.
    \item We provide \emph{comprehensive analysis} through ablations and attention studies, and explore post-hoc explainability by aligning our method to generate natural-language rationales for its boundary predictions.
\end{itemize}

\vspace{-1mm}
\section{Related Work}
\vspace{-1mm}
\paragraph{Video Scene Segmentation}
Segmenting long-form videos into semantically coherent scenes is a long-standing problem in video understanding. Early approaches cast it as an unsupervised clustering task, grouping visually similar shots based on low-level descriptors~\cite{baraldi2015shot}. While label-free, such methods failed to capture narrative-level transitions extending beyond visual continuity. With the emergence of deep learning, hand-crafted features were replaced by CNN-based embeddings~\cite{baraldi2015deep}, and sequential models (e.g., LSTMs) modeled temporal continuity across shots~\cite{huang2020movienet}. The MovieNet dataset~\cite{huang2020movienet} enabled large-scale supervised learning and multimodal integration of subtitles, speech, and character identities~\cite{sadoughi2023mega,rao2020local}.

To further improve representation learning, \textit{ShotCoL}~\cite{chen2021shot} introduced a contrastive self-supervised framework that learns discriminative shot embeddings. Building on this, \textit{BaSSL}~\cite{mun2022boundary} proposed a boundary-aware approach that learns transition cues without annotations. However, both operate on short temporal spans and rely mainly on appearance cues, limiting their ability to capture semantic or narrative context.
More recently, \textit{TranS4mer}~\cite{islam2023efficient} combined self-attention with state-space modeling to efficiently capture long-range dependencies, yet remains an encoder-based classifier without narrative reasoning or explainability.
On the multimodal front, \textit{MEGA}~\cite{sadoughi2023mega} fused visual, subtitle, and screenplay features via multimodal alignment and distillation, but depends on tightly aligned metadata and remains constrained by fixed fusion strategies. Despite these advances, most methods still struggle to capture high-level semantics and generalize beyond the cinematic domain. 

A related task, video chapter segmentation in web videos~\cite{yang2023vidchapters}, defines boundaries semantically without relying on shot units. Models such as \textit{Chapter-LLaMA}~\cite{ventura2025chapter} leverage large language models (LLMs) over transcripts and visual descriptions to segment long-form content, effectively compensating for the lack of high-level semantics that visual-based methods fail to capture; however, as we later show, these models exhibit degraded performance when applied to cinematic materials, likely due to the stronger visual structure of films.

\begin{table}[t!]
\caption{\textbf{High-level comparison of key capabilities across scene segmentation methods.} Scene-VLM supports all 5 capabilities.}
\centering
\footnotesize
\resizebox{\columnwidth}{!}{%
\begin{tabular}{lccccc}
\toprule
\textbf{Method} &
\makecell{\textbf{Sequential}\\\textbf{Prediction}} &
\makecell{\textbf{Confidence}\\\textbf{Scoring}} &
\makecell{\textbf{Explain-}\\\textbf{ability}} &
\makecell{\textbf{Cinematic}} &
\textbf{Chaptering} \\
\midrule
ShotCoL        & \xmark & \cmark & \xmark & \cmark & \xmark \\
BaSSL          & \xmark & \cmark & \xmark & \cmark & \xmark \\
TranS4mer      & \xmark & \cmark & \xmark & \cmark & \xmark \\
MEGA           & \xmark & \cmark & \xmark & \cmark & \xmark \\
Chapter-LLaMA  & \cmark & \xmark & \cmark & \xmark & \cmark \\
\midrule
\textbf{Scene-VLM} (ours)  & \cmark & \cmark & \cmark & \cmark & \cmark \\
\bottomrule
\end{tabular}}
\label{tab:framework_comparison}
\vspace{-4mm}
\end{table}

\cref{tab:framework_comparison} summarizes the capabilities of prior methods relative to ours. Classical video scene segmentation models do not naturally support sequential predictions or explainability and are not designed for chaptering. Chapter-LLaMA, in contrast, is tailored for chaptering and supports sequential generation and explainability, but it neither performs scene segmentation for cinematic content nor provides confidence scores. Our method is the only one that (i) performs \emph{both} tasks competitively, (ii) makes sequential predictions, (iii) exposes confidence scores, and (iv) can produce natural-language rationales for its boundary predictions.

\vspace{-3.5mm}
\paragraph{Vision–Language Models for Video Understanding}
Recent years have witnessed rapid progress in the utilization of vision–language models (VLMs) for various video understanding tasks~\cite{fang2024videollava,wu2023videollm,li2024videochat2,xue2024longvila, zhang2023video, chen2024internvl}. However, video processing remains challenging due to internal context-length limitations. To mitigate this, recent works introduce hierarchical representations and efficient temporal modeling to extend reasoning horizons~\cite{jin2024vidgpt,xue2024longvila, naiman2025lvmaelearninglongvideo}. 
Among open-source models, Qwen2.5-VL~\cite{bai2025qwen2} demonstrates strong multimodal reasoning, making it a viable backbone for structured video understanding.
Despite this emerging potential, to our knowledge our work is the first to explore VLMs for the task of video scene segmentation.

\vspace{-2mm}
\section{Approach}
\vspace{-1mm}
In this section, we outline our proposed approach for leveraging a vision-language model (VLM) to perform video scene segmentation. An overview of our approach is illustrated in~\cref{fig:method_flow}. As depicted, our method involves fine-tuning the VLM to jointly process visual frames, subtitles, and character information in order to identify scene boundaries within a given context window, while associating confidence scores with each shot in a sequential manner. 

\vspace{-1mm}
\subsection{VLM for Video Scene Segmentation}
\vspace{-1mm}
We define a context window $C = \{s_i\}_{i=1}^N$ containing $N$ consecutive shot representations. Each shot representation $s_i$ consists of $K$ sampled frames $\{f_{i,k}\}_{k=1}^K$ along with synchronized subtitles and character information associated with the shot. The context $C$ is then provided to the vision-language model $\mathcal{M}$ along with an instruction prompt $P$, which directs the model to identify shots corresponding to scene boundaries. 
Formally,
\vspace{-1mm}
\begin{equation} \label{eq:vlm}
    Y = \mathcal{M}(C, P),
\end{equation}
where $Y = \{y_j\}_{j=1}^T$ denotes the sequence of output logits produced by the model, and each $y_j \in \mathbb{R}^{|V|}$ represents the predicted token distribution over the vocabulary $V$.
The instruction prompt guides the model to produce its predictions as a sequence of shot-level entries, each following the format \texttt{shot\_id:<id>: Yes/No}.
Following prior work~\cite{islam2023efficient}, we label a shot as positive (\texttt{Yes}) if it marks the end of a new scene. An illustrative example of the output format is shown in~\cref{fig:method_flow}. 
This scheme enables sequential inference, in which the prediction for each shot is conditioned on the predictions of preceding shots.

Since each prediction depends on the surrounding temporal context, shots near the boundaries of the context window tend to yield less reliable predictions. 
To address this, the instruction prompt restricts predictions to a \textit{focus window} centered in the context,
ensuring that each evaluated shot has sufficient temporal evidence from both preceding and following shots.
In practice, the context window typically comprises 20 consecutive shots, and the focus window includes the central 10.

We fine-tune the VLM on an annotated dataset for video scene segmentation (e.g., \cite{huang2020movienet}), where each training sample consists of a context window provided to the model along with the instruction prompt. The model is trained to generate responses $Y$ that align with the target labels (\texttt{Yes}/\texttt{No}) for each shot in the focus window. Training is performed using a next-token prediction loss.
Optionally, the model can be further aligned to generate natural-language explanations for its boundary predictions, as described in Section~\ref{sec:explainability_res}.

\vspace{-1mm}
\subsection{Computing Soft Predictions}
\vspace{-1mm}
\label{sec:confidence_method}
Estimating confidence scores for scene boundary predictions is crucial for controlling the precision–recall trade-off.
Unlike encoder-based approaches to scene boundary detection, where prediction scores can be obtained directly from a dedicated classification head, VLMs generate a sequence of textual output tokens in response to a given instruction prompt. 
Consequently, we propose computing a confidence score for each shot-level decision by probing the logit vectors corresponding to the \texttt{Yes}/\texttt{No} tokens.

Specifically, the probability of a token $t$ at position $j$ in the output sequence is defined as
\begin{equation} \label{eq:softmax}
    p_j(t) = \frac{\exp\!\big(y_j[t]\big)}{\sum_{u \in V} \exp\!\big(y_j[u]\big)}.
\end{equation}
For each shot, we identify the position of its verdict entry in the structured output sequence and compute the probabilities of \texttt{Yes} (scene boundary) and \texttt{No} (continuation).
We denote these as $p_i(\texttt{Yes})$ and $p_i(\texttt{No})$, representing the positive and negative probabilities for shot $i$, respectively.
The confidence score for shot $i$ is then defined as
\begin{equation} \label{eq:conf}
\mathrm{conf}_i = \frac{p_i\!\left(\texttt{Yes}\right)}{p_i\!\left(\texttt{Yes}\right) + p_i\!\left(\texttt{No}\right)}.
\end{equation}
This scheme allows sequential computation of the probability that a given shot contains a scene boundary, conditioned on the multimodal inputs \emph{and} on model predictions for preceding shots. We discuss alternative output formats and additional design considerations in the Appendix.

\vspace{-1mm}
\section{Experiments}
\vspace{-1mm}
We present a comprehensive empirical evaluation of \methodname. We begin by describing the datasets and implementation details (\cref{sec:datasets} and~\cref{sec:impl_details}). We then establish state-of-the-art performance on standard scene segmentation benchmarks (\cref{sec:main_results}). To better understand the source of these gains, we conduct systematic analyses:~\cref{sec:attention_analysis} examines how the model distributes attention across different input modalities and temporal context, and~\cref{sec:abl_study} presents a comprehensive ablation study of the model’s key design choices. In  \cref{sec:chaptering} we evaluate generalization to the related video chaptering task. Finally, in ~\cref{sec:explainability_res} we explore aligning our framework to generate coherent verbal explanations for its boundary decisions through targeted fine-tuning on a small set of annotated samples. More details and additional experiments are provided in the Appendix.

\vspace{-1mm}
\subsection{Datasets}
\vspace{-1mm}
\label{sec:datasets}
\paragraph{MovieNet-318.} This dataset is a subset of the MovieNet dataset~\cite{huang2020movienet} for scene segmentation in cinematic content. It contains 318 movies with shot-level annotations and scene boundary labels, split into 190 for training, 64 for validation, and 64 for testing. On average, each movie contains around 1000 shots, with each shot annotated with a binary label indicating whether it marks a scene boundary.

\vspace{-3.5mm}
\paragraph{BBC Planet Earth.} This dataset~\cite{baraldi2015deep} is a standard out-of-distribution benchmark for scene segmentation, consisting of 10 documentary episodes from the Planet Earth series. Each episode averages 50 minutes in duration, with 670 scenes and 4.9K shots in total. Unlike cinematic content, these episodes feature documentary-style narration, non-fiction pacing, and highly diverse visual domains (wildlife, landscapes, climate footage).

\vspace{-3.5mm}
\paragraph{VidChapters-7M.} This dataset~\cite{yang2023vidchapters} addresses the task of segmenting web videos into chapters with timestamped titles. It aggregates videos and user-defined annotations from YouTube. Since the full dataset is not publicly released, we follow~\cite{ventura2025chapter} and use a reproducible subset consisting of 1000 training samples and 300 evaluation videos.

\vspace{-1mm}
\subsection{Implementation Details}
\vspace{-1mm}
\label{sec:impl_details}
We use Qwen2.5-VL-7B~\cite{bai2025qwen2} as our base model. To construct the multimodal shot representations, we: (1) segment videos into shots using standard methods~\cite{pal2015video,boreczky1996comparison} or use provided shot annotations; (2) extract per-shot transcripts using Whisper~\cite{radford2023robust} or use provided subtitles; and (3) add per-shot metadata (e.g., actor identities) when available. Unless stated otherwise, we use a context window of 20 shots, a focus window of 10 shots, and sample 3 frames per shot in all experiments. In addition, we overlay a small visual identifier (shot-ID marker) on each frame to help the model associate visual content with the corresponding shot references in the textual input sequence. Additional training, inference and input related details are provided in the Appendix.

\vspace{-1mm}
\subsection{Scene Segmentation Results}
\vspace{-1mm}
\label{sec:main_results}

\paragraph{MovieNet-318.} We train on MovieNet-318 and evaluate on the test split. Following prior work we report F1 and Average Precision (AP) scores, and compare against leading approaches. As depicted in~\cref{tab:movinet_results_main}, \methodname~ establishes a new state of the art on MovieNet-318, substantially outperforming previous work. In particular, we achieve a gain of $+6.8$ in F1 and $+8.2$ in AP over MEGA \citep{sadoughi2023mega}, and $+13.7$ in F1 and $+6.0$ in AP over TranS4mer \citep{islam2023efficient}. We also compare against Chapter-LLaMA~\citep{ventura2025chapter}, a recent LLM-based video chaptering method that we adapt to the task by generating per-shot captions (see Appendix for details). While Chapter-LLaMA performs well on web videos, it underperforms on cinematic content, likely due to the stronger visual structure of films requiring direct visual processing rather than caption-based reasoning.

\begin{table}[t!]
  \centering
  \footnotesize
  \caption{\textbf{Results on MovieNet-318.} Scene segmentation performance on the MovieNet-318 dataset.}
    \begin{tabular}{p{4.5cm}cc}
      \toprule
      \textbf{Method} & \textbf{F1}~$\uparrow$ & \textbf{AP}~$\uparrow$ \\
      \midrule
      Chapter-LLaMA~\cite{ventura2025chapter} & 38.6                & 41.5 \\
      LGSS~\cite{rao2020local}              & \textemdash           & 47.1 \\
      ShotCoL~\cite{chen2021shot}           & \textemdash           & 53.4 \\
      BaSSL~\cite{mun2022boundary}          & 47.0                  & 57.4 \\
      MEGA~\cite{sadoughi2023mega}          & \underline{55.3}      & 58.6 \\
      TranS4mer~\cite{islam2023efficient}   & 48.4                  & \underline{60.8} \\
      \midrule
      \textbf{Scene-VLM (ours)}                  & \textbf{62.1}         & \textbf{66.8} \\
      \bottomrule
    \end{tabular}
    \vspace{-4mm}
  \label{tab:movinet_results_main}
\end{table}

\vspace{-3.5mm}
\paragraph{BBC Planet Earth.}
Following prior work~\citep{mun2022boundary,bertasius2021space,islam2023efficient}, we evaluate \emph{zero-shot} on BBC after training on MovieNet-318 to assess cross-domain generalization. Since prior work does not report F1 on BBC, we report only AP. As seen in~\cref{tab:bbc_ap_scores}, \methodname~sets a new zero-shot state of the art on BBC, outperforming the previous leading method TranS4mer ~\citep{islam2023efficient} by +2.2 AP.

\begin{table}[htpb]
    \vspace{-2mm}
  \centering
  \footnotesize

  \begin{tabular}{p{4.5cm}c}
    \toprule
    \textbf{Method} & \textbf{AP}~$\uparrow$ \\
    \midrule
    BaSSL~\cite{mun2022boundary} & 40.0 \\
    TimeSformer~\cite{bertasius2021space} & 42.2 \\
    TranS4mer~\cite{islam2023efficient} & \underline{43.6} \\
    \midrule
    \textbf{Scene-VLM (ours)} & \textbf{45.8} \\
    \bottomrule
  \end{tabular}
    \caption{\textbf{Results on BBC Planet Earth.} Zero-shot scene segmentation performance on the BBC Planet Earth dataset.}
  \label{tab:bbc_ap_scores}
  \vspace{-4mm}
\end{table}

\vspace{-2mm}
\subsection{Attention Analysis}
\vspace{-1mm}
\label{sec:attention_analysis}
Each output token generated by the VLM, particularly the shot-level verdict token (\texttt{Yes}/\texttt{No} for a given shot), attends to all preceding output tokens as well as to the input tokens provided to the model.
In this section, we analyze how each input component contributes to the model’s decisions by examining the attention allocated to the input modalities and the previously generated output tokens.

We denote by $A_{ij}$ the attention weight between an output token $i$ and any preceding token $j$ (from either the input or previously generated outputs). We then compute the contribution of each component by averaging attention values across all layers and attention heads. For example, to obtain aggregate attention between a given shot-prediction token at position $i$ and the set of visual input tokens $\mathcal{V}$, we compute $\sum_{j \in \mathcal{V}} A_{ij}$, averaged over all layers and attention heads.

We begin by analyzing the contribution of each component to the model’s predictions by aggregating attention values across all shots for each input modality, and for the preceding output tokens.~\cref{fig:modality-attention-two-panel} (left) shows the aggregated attention distribution across the input modalities: visual, subtitles, and actor ID. As observed, visual tokens receive the largest share of attention, indicating their dominant role in predicting scene transitions. The preceding output tokens also receive substantial attention, highlighting the importance of sequential inference enabled by the method design.

In~\cref{fig:modality-attention-two-panel} (right), we present the attention distribution after normalizing each modality by its token count. For example, for the visual modality, we compute $\frac{1}{|\mathcal{V}|}\sum_{j \in \mathcal{V}} A_{ij}$ to normalize its aggregate attention.
This normalization allows us to assess the relative importance of each modality while eliminating biases introduced by differences in token counts.
We exclude the output tokens from this normalization since, being few but densely connected, they would otherwise dominate the average.
As shown, after normalization, subtitle and actor tokens receive attention comparable to visual tokens, indicating that these modalities provide valuable cues for identifying scene transitions in video.

\usetikzlibrary{positioning}
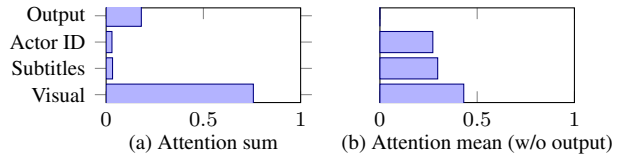
\begin{figure}[t!]
\vspace{3mm}
\centering
\begin{tikzpicture}
\begin{groupplot}[
    group style={group size=2 by 1, horizontal sep=30pt},
    width=0.50\linewidth,
    height=4.5cm,
    xbar,
    xmin=0, xmax=1.0,
    symbolic y coords={Visual, Subtitles, Actor~ID, Output},
    ytick=data,
    yticklabel style={font=\footnotesize},
    xticklabel style={font=\footnotesize},
    xtick={0,0.5,1.0},
    xlabel={}, 
    y=0.35cm,
]

\nextgroupplot
\addplot+[bar width=8pt, fill=blue!30, draw=blue!50!black] coordinates {
(0.757,Visual)
(0.033,Subtitles)
(0.030,Actor~ID)
(0.181,Output)
};

\nextgroupplot[
    yticklabels={,, ,},
    ytick style={draw=none},
]
\addplot+[bar width=8pt, fill=blue!30, draw=blue!50!black] coordinates {
(0.431,Visual)
(0.297,Subtitles)
(0.272,Actor~ID)
(0.000,Output)
};

\end{groupplot}

\node[font=\footnotesize] at ($(group c1r1.south) + (0,-0.55cm)$) {(a) Attention sum };
\node[font=\footnotesize] at ($(group c2r1.south) + (0,-0.55cm)$) {(b) Attention mean (w/o output)};

\end{tikzpicture}

\vspace{-2mm}
\caption{\textbf{Attention by modality.} Visualization of attention distribution across input modalities (visual, subtitles, and actor IDs) as well as preceding output shot predictions. 
\textbf{(a)} Summed attention reveals strong visual dominance and high dependency on prior output tokens. 
\textbf{(b)} Averaged (length-normalized) attention highlights that subtitles and actor IDs contribute comparably to visual tokens.}
\label{fig:modality-attention-two-panel}
\vspace{-6mm}
\end{figure}

\begin{figure*}[t]
\centering

\begin{subfigure}[t]{0.32\textwidth}
  \centering
  \includegraphics[width=\linewidth]{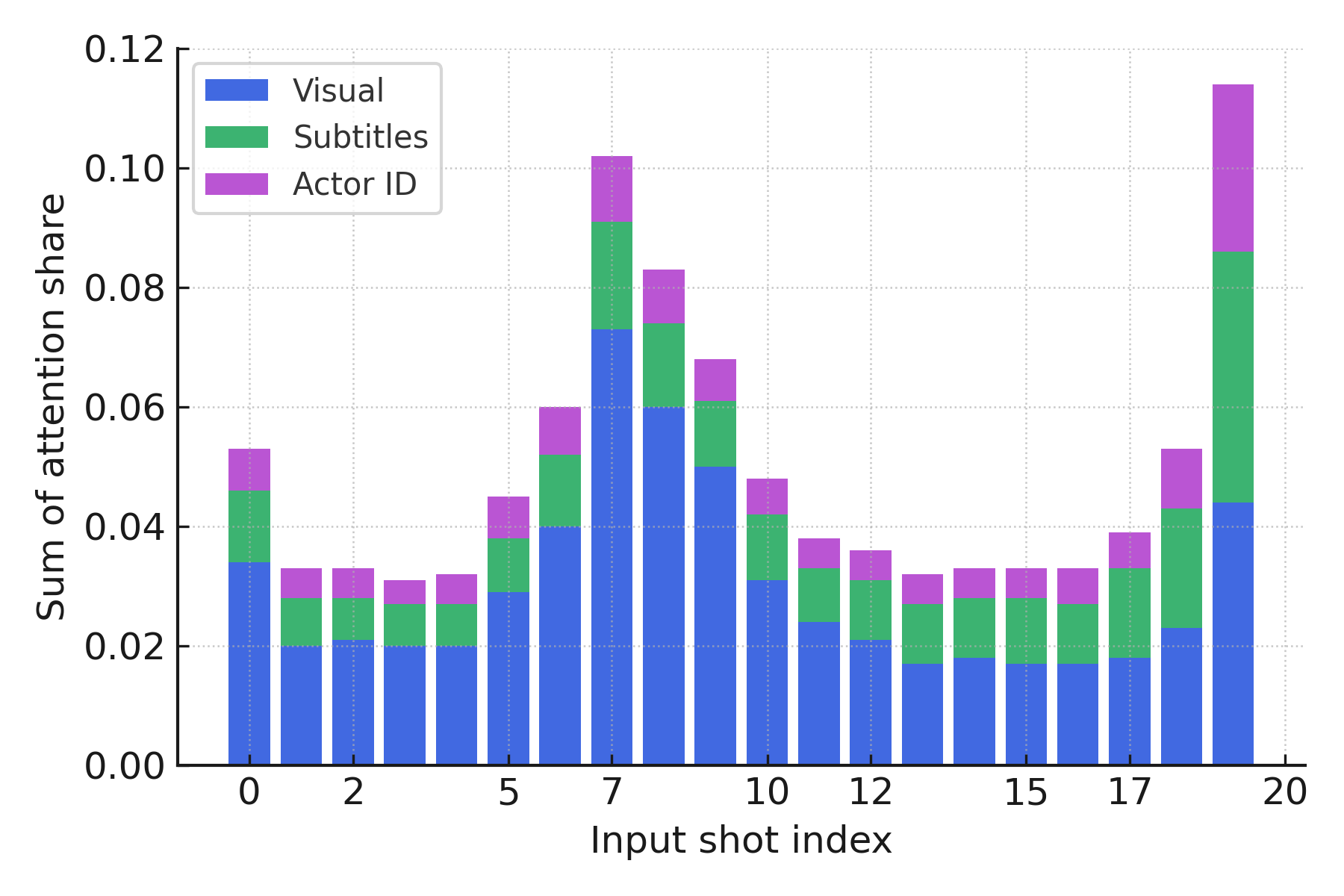}
  \caption{Shot 7 prediction}
  \label{fig:attn_shot7}
\end{subfigure}
\hfill
\begin{subfigure}[t]{0.32\textwidth}
  \centering
  \includegraphics[width=\linewidth]{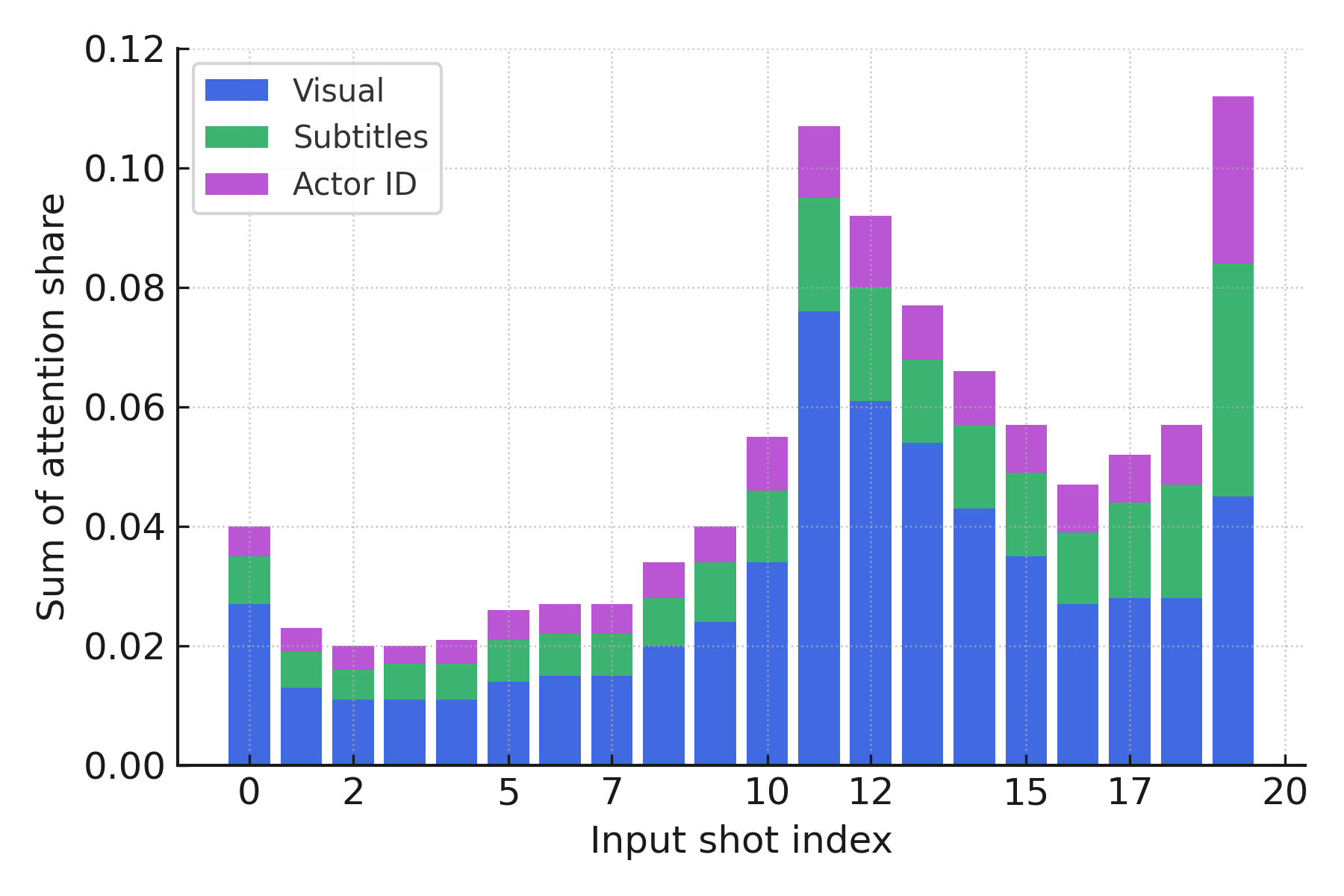}
  \caption{Shot 11 prediction}
  \label{fig:attn_shot11}
\end{subfigure}
\hfill
\begin{subfigure}[t]{0.32\textwidth}
  \centering
  \includegraphics[width=\linewidth]{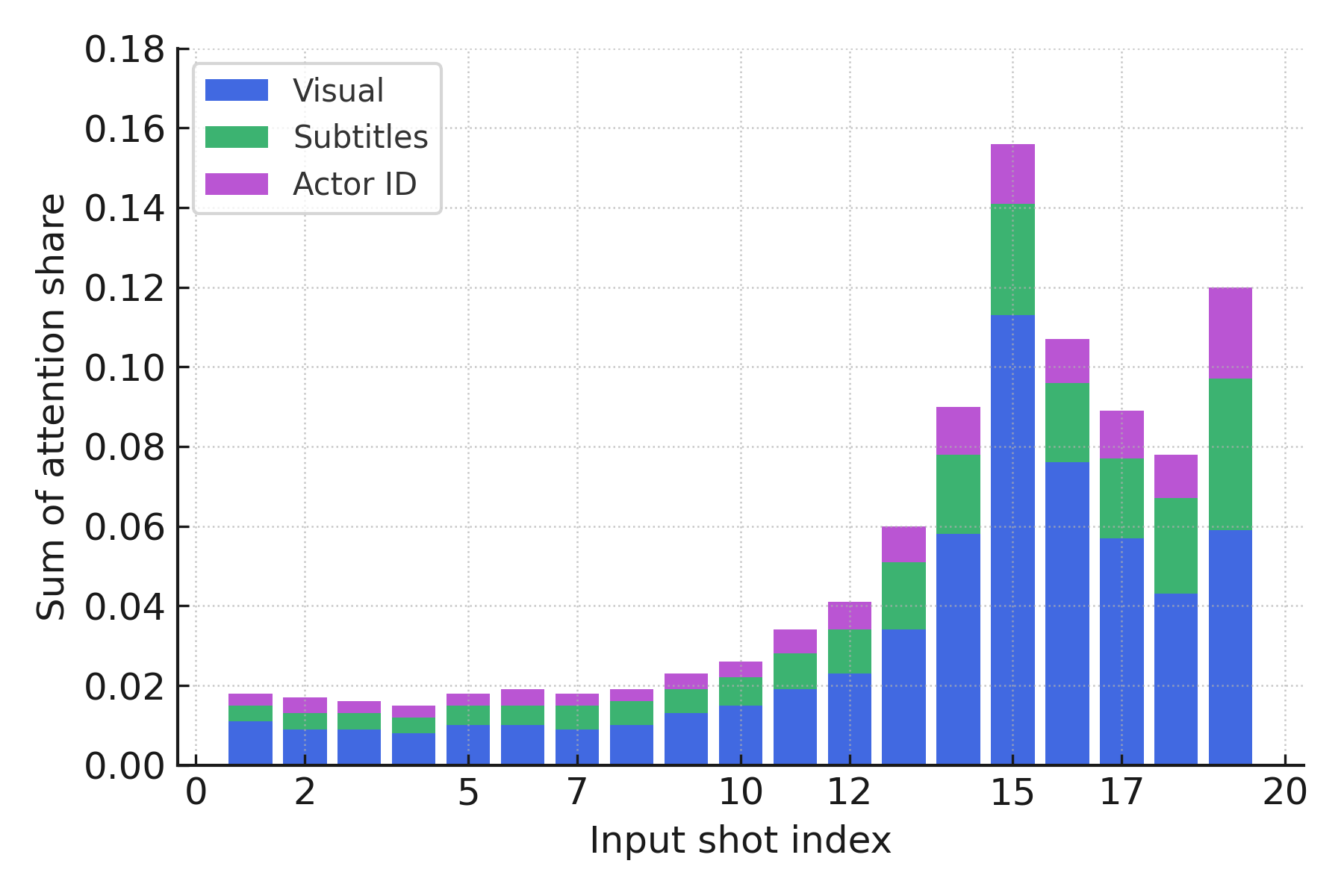}
  \caption{Shot 15 prediction}
  \label{fig:attn_shot15}
\end{subfigure}

\caption{\textbf{Attention distributions across shots and modalities.} 
Figures show modality-level stacked attention shares for three shot predictions: 
(a) \textbf{Shot 7}, (b) \textbf{Shot 11}, and (c) \textbf{Shot 15}. 
Attention is computed between the output token of the corresponding shot prediction and the input tokens for all shots.
Each bar represents the relative attention of the visual, subtitle, and actor-ID input modalities per shot.}
\label{fig:three_attention_subfigs}
\vspace{-5mm}
\end{figure*}

Next, we analyze how each shot prediction attends to individual shots in the input sequence. We fix a specific shot index and average the attention over multiple samples where a positive scene transition occurs at that index. For each shot, we aggregated attention between the output token corresponding to that shot and the input tokens of each shot, and also examine the relative attention across the visual, subtitle, and actor-ID modalities.
In~\cref{fig:three_attention_subfigs}, we show the resulting attention distributions for three output predictions, shots 7, 11, and 15, each averaged over 30 context samples sharing the same positive transition index.
Interestingly, as observed particularly for shots 11 (b) and 15 (c), the output token corresponding to the shot prediction attends more strongly to the subsequent shots in the input sequence than to the preceding ones. 
We hypothesize that this behavior arises because the prediction for a given shot already encodes information about the preceding shots through the previously generated output tokens. Consequently, the model allocates less attention to earlier shots it has already “seen” and more to subsequent ones, which provide additional context necessary for identifying the scene transition. This observation supports our findings in Sec.~\ref{sec:focus-and-output}, reinforcing that sequential prediction plays a crucial role. The model appears to \textit{trust} its previous predictions, allocating less attention to shots it has already processed and focusing instead on future inputs that may refine or confirm its ongoing decision.

Finally, we observe a high peak of attention at the first and last input shots. We hypothesize that this behavior helps the model identify the temporal boundaries of the input sequence more easily.

\vspace{-1mm}
\subsection{Ablation Study}
\vspace{-1mm}
\label{sec:abl_study}
We conduct systematic ablations to validate our design choices and quantify the contribution of
different components. We examine (1) the relative importance of visual, textual, and metadata inputs, (2) the necessity of the context-focus window design for position-invariant predictions and the impact of different window sizes, (3) the impact of the number of key frames per shot, and (4) model size effects. Ablations are conducted on MovieNet-318.

\subsubsection{Input Components Analysis}
\label{sec:input_ablt}
To understand which input modalities drive performance, we explore the contribution of four components:
(1) Visual key frames, (2) shot-ID markers---visual tags linking each frame to its shot identifier in the prompt (see Appendix), (3)
subtitles, and (4) actor-IDs. We perform two complementary ablations: (i) remove one component at a time
from the full model, and (ii) keep a single component and remove the rest.

The results shown in~\cref{tab:input_components_ablation} reveal clear performance hierarchies
across components. Visual removal causes catastrophic failure (F1: 62.1 → 32.0), establishing vision as
the primary boundary cue. However, other components provide meaningful contributions: Shot-ID
removal drops performance by 1.3 F1 points, indicating that temporal anchoring still matters beyond raw
visuals, while subtitle and actor-ID removal each cause approximately 1 point drops, showing these components contribute complementary signals.

The isolation experiments further confirm this hierarchy. Visual-only achieves strong performance (58.6
F1), showing that many scene boundaries have clear visual signatures. In contrast, subtitle-only and
actor-only configurations degrade sharply, implying that textual signals alone lack sufficient boundary
evidence.

\begin{table}[htbp]
  \vspace{-2mm}
  \centering
  \footnotesize
  \caption{\textbf{Input component ablation.} Top section shows removal of individual components; bottom section shows performance with only single components.}
  \resizebox{\columnwidth}{!}{%
  \begin{tabular}{cccc|cc}
    \toprule
    \textbf{Visual} & \textbf{Shot-ID} & \textbf{Subtitles} & \textbf{Actor-ID} & \textbf{F1}~$\uparrow$ & \textbf{AP}~$\uparrow$ \\
    \midrule
    \cmark & \cmark & \cmark & \cmark & 62.1 & 66.8 \\
    \xmark & \xmark & \cmark & \cmark & 32.0 & 34.7 \\
    \cmark & \xmark & \cmark & \cmark & 60.8 & 64.1 \\
    \cmark & \cmark & \xmark & \cmark & 61.1 & 62.2 \\
    \cmark & \cmark & \cmark & \xmark & 61.3 & 62.0 \\
    \midrule
    \cmark & \xmark & \xmark & \xmark & 58.6 & 61.4 \\
    \xmark & \xmark & \cmark & \xmark & 31.5 & 33.2 \\
    \xmark & \xmark & \xmark & \cmark & 24.8 & 28.6 \\
    \bottomrule
  \end{tabular}}
  \vspace{-4mm}
  \label{tab:input_components_ablation}
\end{table}

\subsubsection{Context-Focus Window and Sequential Prediction}
\label{sec:focus-and-output}

We next study two aspects of our sequential prediction approach: whether context margins prevent performance degradation at edge positions, and how different context-focus window sizes affect performance while demonstrating the benefits of sequential over point-wise prediction.

\vspace{-5mm}
\paragraph{Performance Degradation at Sequence Edges.}

To validate the necessity of our focus mechanism in preventing performance degradation near sequence boundaries, we analyze per-position performance across 10 sequential predictions. We compare two settings: with focus (predicting only for the central 10 shots within a 20-shot context window) versus without focus (predicting for all 10 shots without surrounding context).

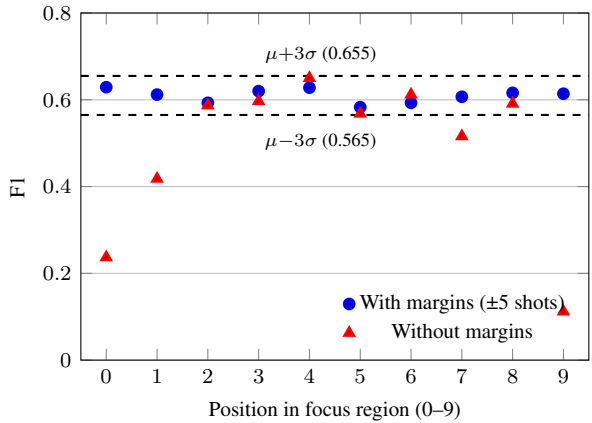
\begin{figure}[htpb]
\vspace{-2mm}
\centering
\begin{tikzpicture}
\begin{axis}[
    width=\columnwidth,
    height=6.2cm,
    xmin=-0.5, xmax=9.5,
    ymin=0, ymax=0.8,
    xlabel={Position in focus region (0--9)},
    ylabel={F1},
    xtick={0,1,2,3,4,5,6,7,8,9},
    ymajorgrids=true,
    legend style={at={(0.98,0.02)}, anchor=south east, draw=none, fill=none, font=\footnotesize},
    tick label style={font=\footnotesize},
    label style={font=\footnotesize}
]

\addplot+[only marks, mark=*, mark size=2pt, thick]
coordinates {
(0,0.629) (1,0.612) (2,0.593) (3,0.620) (4,0.628)
(5,0.583) (6,0.593) (7,0.607) (8,0.616) (9,0.614)
};
\addlegendentry{With margins (±5 shots)}

\addplot+[only marks, mark=triangle*, mark size=2.2pt, thick]
coordinates {
(0,0.237) (1,0.418) (2,0.587) (3,0.597) (4,0.650)
(5,0.568) (6,0.612) (7,0.516) (8,0.591) (9,0.112)
};
\addlegendentry{Without margins}

\addplot[black, dashed, line width=0.8pt] coordinates {(-0.5,0.655) (9.5,0.655)};
\addplot[black, dashed, line width=0.8pt] coordinates {(-0.5,0.565) (9.5,0.565)};
\node[anchor=south east, font=\scriptsize] at (axis cs:5.5,0.665) {$\mu{+}3\sigma$ (0.655)};
\node[anchor=north east, font=\scriptsize] at (axis cs:5.5,0.545) {$\mu{-}3\sigma$ (0.565)};

\end{axis}
\end{tikzpicture}
\vspace{-3mm}
\caption{\textbf{Focus mechanism prevents edge degradation.} Performance collapses at boundaries without focus mechanism (red) but remains stable with focus mechanism (blue).}
\label{fig:position_f1_outliers}
\vspace{-5mm}
\end{figure}

\cref{fig:position_f1_outliers} shows the per-position F1 scores for both configurations. \emph{Without the focus mechanism} (red triangles), performance collapses dramatically at sequence boundaries, with edge positions showing severe degradation compared to central positions. \emph{With the focus mechanism} (blue circles), performance remains consistent across all positions, with no outliers beyond 3 standard deviations from the mean. %
This demonstrates that temporal context margins are essential for position-invariant performance.

\vspace{-3mm}
\paragraph{Context \& Focus Window Sizes.}

We systematically ablate context and focus window sizes by testing $l_{\text{context}}\!\in\!\{20,10,5\}$ and $l_{\text{focus}}\!\in\!\{20,10,5,1\}$. 
Results in~\cref{tab:ctx-focus-ablation} show two consistent trends. First, \textbf{sequential prediction matters}: for any context window size, reducing focus to a single shot ($l_{\text{focus}}{=}1$) consistently degrades
performance, demonstrating that the model benefits from causally linked predictions across multiple
shots. Second, \textbf{longer context helps}: increasing context window size (while keeping margins) reliably improves performance, indicating that broader temporal evidence aids boundary disambiguation. Moreover, as previously discussed, removing temporal margins causes edge degradation, highlighting the need for context padding to ensure position-invariant predictions.

\vspace{-1mm}
\begin{table}[htbp]
  \centering
  \caption{\textbf{Window size ablation.} Each column shows context $\&$ focus window sizes with corresponding F1 scores.}
  \vspace{-2mm}
  \small
  \resizebox{\columnwidth}{!}{%
  \begin{tabular}{cccc|ccc|cc}
    \toprule
     20 \& 20 & 20 \& 10 & 20 \& 5 & 20 \& 1 & 10 \& 10 & 10 \& 5 & 10 \& 1 & 5 \& 5 & 5 \& 1 \\
    \midrule
     59.7 & \textbf{62.1} & 61.9 & 60.1 & 58.4 & \textbf{60.9} & 59.4 & \textbf{55.8} & 54.9 \\
    \bottomrule
  \end{tabular}}
  \label{tab:ctx-focus-ablation}
  \vspace{-4mm}
\end{table}

\subsubsection{Number of Frames per Shot}
\label{sec:num-frames-per-shot}

We study the effect of frames-per-shot ($K$). While higher $K$ can capture more intra-shot dynamics, it inflates token count. \cref{tab:keyframe_ablation} shows results for $K\!\in\!\{1,2,3\}$. Performance improves modestly but consistently as $K$ increases. The modest gains suggest that for scene segmentation, a single representative frame suffices for most shots, though additional frames provide complementary evidence, presumably for shots with significant intra-shot motion or visual complexity. 
See further computational analysis in the Appendix.

\begin{table}[t]
\centering
\footnotesize
\begin{minipage}[t]{0.47\columnwidth}
    \caption{Frames per shot}
        \vspace{-2mm}
    \centering
    \begin{tabular}{l|cc}
        \toprule
        \textbf{$K$} & \textbf{F1} & \textbf{AP} \\
        \midrule
        1 & 61.8 & 65.3  \\
        2 & 61.9 & 65.2 \\
        3 & \textbf{62.1} & \textbf{66.8} \\
        \bottomrule
    \end{tabular}
    \label{tab:keyframe_ablation}
\end{minipage}
\hfill
\begin{minipage}[t]{0.47\columnwidth}
    \caption{Model size}
    \vspace{-2mm}
    \centering
    \begin{tabular}{l|cc}
        \toprule
        \textbf{Params}~$\uparrow$ & \textbf{F1} & \textbf{AP} \\
        \midrule
        1.5B & 55.9 & 58.7 \\
        3B   & 59.6 & 62.8 \\
        7B   & \textbf{62.1} & \textbf{66.8} \\
        \bottomrule
    \end{tabular}
    \label{tab:model_size_ablation}
\end{minipage}
\vspace{-4mm}
\end{table}

\subsubsection{Model Size}

To understand how model capacity affects performance, we evaluate three model sizes - \emph{1.5B}, \emph{3B}, and \emph{7B} parameters. \cref{tab:model_size_ablation} shows consistent, monotonic improvements as model size
increases: scaling from 1.5B to 3B yields gains of +3.7 F1 and +4.1 AP, while further scaling to 7B adds
+2.5 F1 and +4.0 AP. Notably, the gains remain substantial even at the largest scale, suggesting that
further scaling may continue to improve performance. These results provide empirical evidence that scene
segmentation benefits from increased model capacity, consistent with broader scaling trends observed in
vision-language models.

\vspace{-1mm}
\subsection{Adaptation to Video Chaptering}
\vspace{-1mm}
\label{sec:chaptering}

To assess generalization beyond cinematic scene segmentation, we evaluate on the related task of video chaptering~\citep{yang2023vidchapters, ventura2025chapter}. Unlike scene segmentation, this task targets web videos,  where chapter boundaries are defined semantically and do not necessarily align with shot boundaries. The task also requires predicting \emph{both} chapter boundary timestamps and a descriptive title for each chapter.

\vspace{-3mm}
\paragraph{Experimental Setup.}
We adapt our framework with a minimal change: instead of emitting binary boundary labels, the model outputs boundary times with corresponding titles, following the format \texttt{hh:mm:ss - Title}~\cite{ventura2025chapter}. We then evaluate on a subset of the VidChapters-7M corpus~\cite{yang2023vidchapters,ventura2025chapter}, where content creators provide timestamped chapter titles. The primary baseline is Chapter-LLaMA~\cite{ventura2025chapter}, which uses a LLaMA-3.1 backbone \cite{grattafiori2024llama}. To isolate methodology from backbone effects, we also compare against a variant of Chapter-LLaMA that uses a Qwen2.5-VL~\cite{bai2025qwen2} backbone of comparable size, keeping the rest of the pipeline identical. We report F1, temporal IoU (tIoU), SODA, and CIDEr, jointly reflecting boundary accuracy, temporal alignment, and title quality.

\vspace{-3mm}
\paragraph{Results and Analysis.}
As shown in~\cref{tab:chaptering_results}, replacing the backbone in Chapter-LLaMA yields a notable performance drop, indicating that a direct backbone substitution is non-trivial. However, under the matched Qwen backbone, our method outperforms the adapted baseline across all metrics.
Although the original Chapter-LLaMA with its native LLaMA backbone remains strongest in absolute terms, our framework demonstrates clear methodological gains when controlling for the underlying VLM.

\vspace{-2mm}
We attribute these gains to fundamental differences in how visual information is processed. Chapter-LLaMA is a text-only approach: it first generates text captions from the visual signals, then processes these captions alongside speech transcripts. Our approach instead learns directly from raw visual frames jointly with transcripts, avoiding the intermediate captioning step. We believe this end-to-end multimodal grounding enables our model to discover which visual cues are truly predictive for boundary decisions, rather than relying on caption representations that may lose salient visual information.

\begin{table}[t]
  \caption{\textbf{Results on Video Chaptering.}  Under matched backbones \methodname~ outperforms Chapter-LLaMA across all metrics. Chapter-LLaMA with its original LLaMA backbone is listed \textcolor{gray}{in gray} for reference.}
  \vspace{-3mm}
  \centering
  \footnotesize
  \resizebox{\columnwidth}{!}{%
  \begin{tabular}{l l cccc}
    \toprule
    \textbf{Method} & \textbf{Backbone} & \textbf{F1}~$\uparrow$ & \textbf{tIoU}~$\uparrow$ & \textbf{SODA}~$\uparrow$ & \textbf{CIDEr}~$\uparrow$ \\
    \midrule
    \textcolor{gray}{Chapter-LLaMA} 
      & \textcolor{gray}{LLaMA~3.1-8B} 
      & \textcolor{gray}{42.6} 
      & \textcolor{gray}{70.6} 
      & \textcolor{gray}{16.4} 
      & \textcolor{gray}{82.4} \\
    \midrule
    Chapter-LLaMA & Qwen2.5-VL-7B & 28.4 & 59.5 & 10.1 & 45.5 \\
    \textbf{\methodname~ (ours)} & Qwen2.5-VL-7B & \bfseries 32.2 & \bfseries 63.9 & \bfseries 10.6 & \bfseries 52.2 \\
    \bottomrule
  \end{tabular}}
  \vspace{-1mm}
  \label{tab:chaptering_results}
  \vspace{-4mm}
\end{table}

\vspace{-2mm}
\subsection{Post-hoc Explainability for Scene Segmentation}
\vspace{-1mm}
\label{sec:explainability_res}

Scene boundaries in practical workflows are often reviewed by human editors. Providing natural-language rationales alongside boundary proposals can greatly improve usability, allowing editors to judge whether the model’s rationale aligns with narrative intent. Yet, all prior methods for video scene segmentation offer no such capability. Leveraging the verbal capability of the VLM, we extend \methodname~to produce concise textual rationales for its boundary decisions. To our knowledge, this is the first demonstration of rationale generation for video scene segmentation. 

Our initial approach was simple: we modified the prompt of our fine-tuned model (trained only on boundary detection) to request an explanation for each predicted boundary. However, this naive prompting strategy proved inadequate, as it led to frequent formatting errors and hallucinations, rendering the explanations unreliable for practical use.

\vspace{-5mm}
\paragraph{Alignment via Minimal Supervision.}
Since prompting alone proved insufficient, we turned to explicit supervision. However, since no large-scale annotated explanation datasets exist for this task, we instead set to explore whether the model's generation behavior could be \emph{aligned} toward producing well-structured, grounded explanations using \emph{minimal targeted supervision}. To test this, we curated a small set of 35 human-annotated examples, each pairing a boundary with a short rationale describing the narrative transition (e.g., location, time, characters, dialogue topic). An additional fine-tuning stage on this small set yielded an augmented model, \textbf{Scene-VLM + Explain}, capable of producing coherent, structured explanations. \cref{fig:explainability-example} shows this model's capability. More examples are in the Appendix.

\begin{figure}[htbp]
    \centering
    \includegraphics[width=0.77\linewidth]{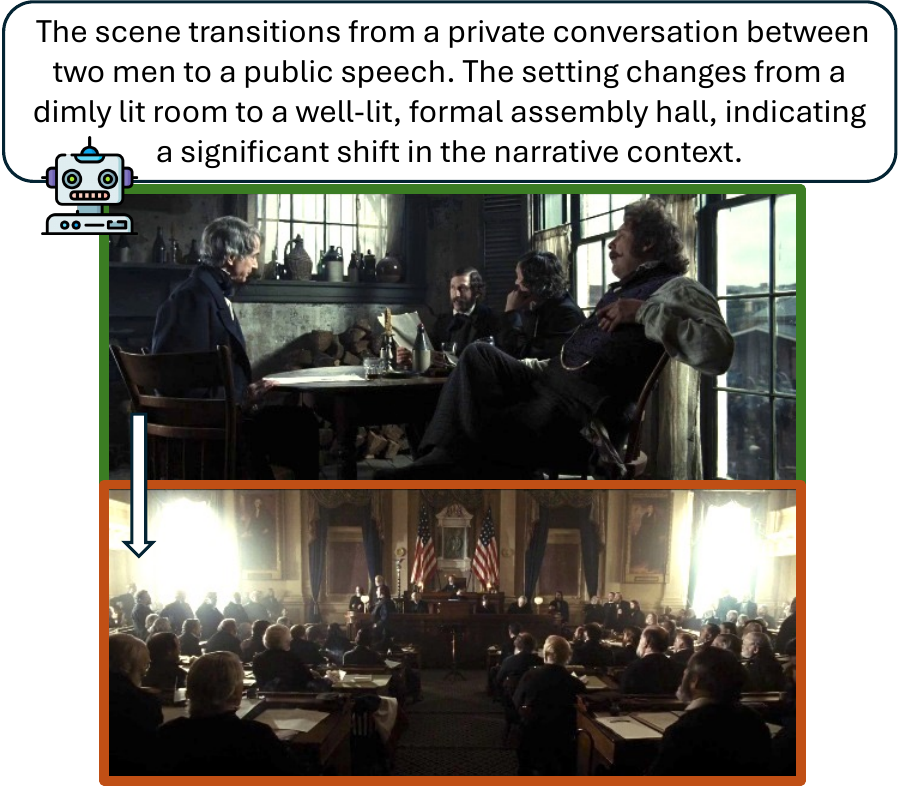}
    \vspace{-2mm}
    \caption{\textbf{Scene transition explanation example.} Scene-VLM + Explain proposes a boundary in \emph{Lincoln} and provides a brief rationale grounded in visual changes, dialogue, and character presence.}
    \label{fig:explainability-example}
    \vspace{-3mm}
\end{figure}

\begin{table}[t]
  \caption{\textbf{Explainability evaluation.} Comparison of explanation quality between Scene-VLM and Scene-VLM + Explain on 30 randomly sampled transitions. The explanation-supervised variant eliminates all formatting errors and hallucinations.}
  \vspace{-1mm}
  \centering
  \footnotesize
  \begin{tabular}{lcc}
    \toprule
    \textbf{Model} & \textbf{Parsing failures} $\downarrow$ & \textbf{Hallucinations} $\downarrow$ \\
    \midrule
    Scene-VLM                & 22 / 30  & 14 / 30 \\
    Scene-VLM + Explain      & \textbf{0 / 30} & \textbf{0 / 30} \\
    \bottomrule
  \end{tabular}
    \vspace{-4mm}
  \label{tab:exp_explainability_format_hallu}
  \vspace{-2mm}
\end{table}

\vspace{-4mm}
\paragraph{Evaluation.}
To assess explanation quality, we conduct a small-scale quantitative probe on 30 randomly sampled transitions from four test movies. We evaluate two criteria: (i)~\textbf{parsing failures}, whether the model produces well-formed, parseable output in the required format, and (ii)~\textbf{hallucinations}, whether rationales contain factually incorrect details or off-task content. As shown in Tab.~\ref{tab:exp_explainability_format_hallu}, the base \methodname{} model (without explanation training) exhibits substantial failure rates. In contrast, Scene-VLM + Explain achieves zero failures on both metrics, demonstrating that minimal targeted supervision is sufficient to align the model toward reliable, well-structured explanations.

\vspace{-1mm}
\section{Conclusion}
\vspace{-1mm}
We introduced \methodname, the first VLM-based framework for video scene segmentation. Our approach addresses key limitations of prior methods through a structured multimodal shot representation, sequential predictions based on a context-focus window design and the ability to generate natural-language rationales for boundary decisions. We further propose a confidence prediction scheme that provides flexible precision-recall trade-offs, a capability typically reserved for encoder-based methods. \methodname~achieves state-of-the-art results on MovieNet and demonstrates strong generalization to BBC Planet Earth and to the video chaptering task. One limitation is the structured \texttt{Yes}/\texttt{No} output format, which trades generative flexibility for reliable confidence extraction; future work could explore more flexible schemes. Looking ahead, we plan to explore reinforcement learning for integrating explicit reasoning into the prediction process, jointly improving both accuracy and interpretability.

\clearpage
{
    \small
    \bibliographystyle{ieeenat_fullname}
    \bibliography{main}
}

\clearpage
\maketitlesupplementary
\setcounter{section}{0}
\renewcommand{\thesection}{\Alph{section}}
\usetikzlibrary{positioning,fit,arrows.meta,shadows.blur}

\section{Additional Details}
\label{sup:additional_details}

\subsection{Method}
\subsubsection{Prompt Structure and Output Format}
\begin{figure*}[t!]
    \centering
    \includegraphics[width=1\linewidth]{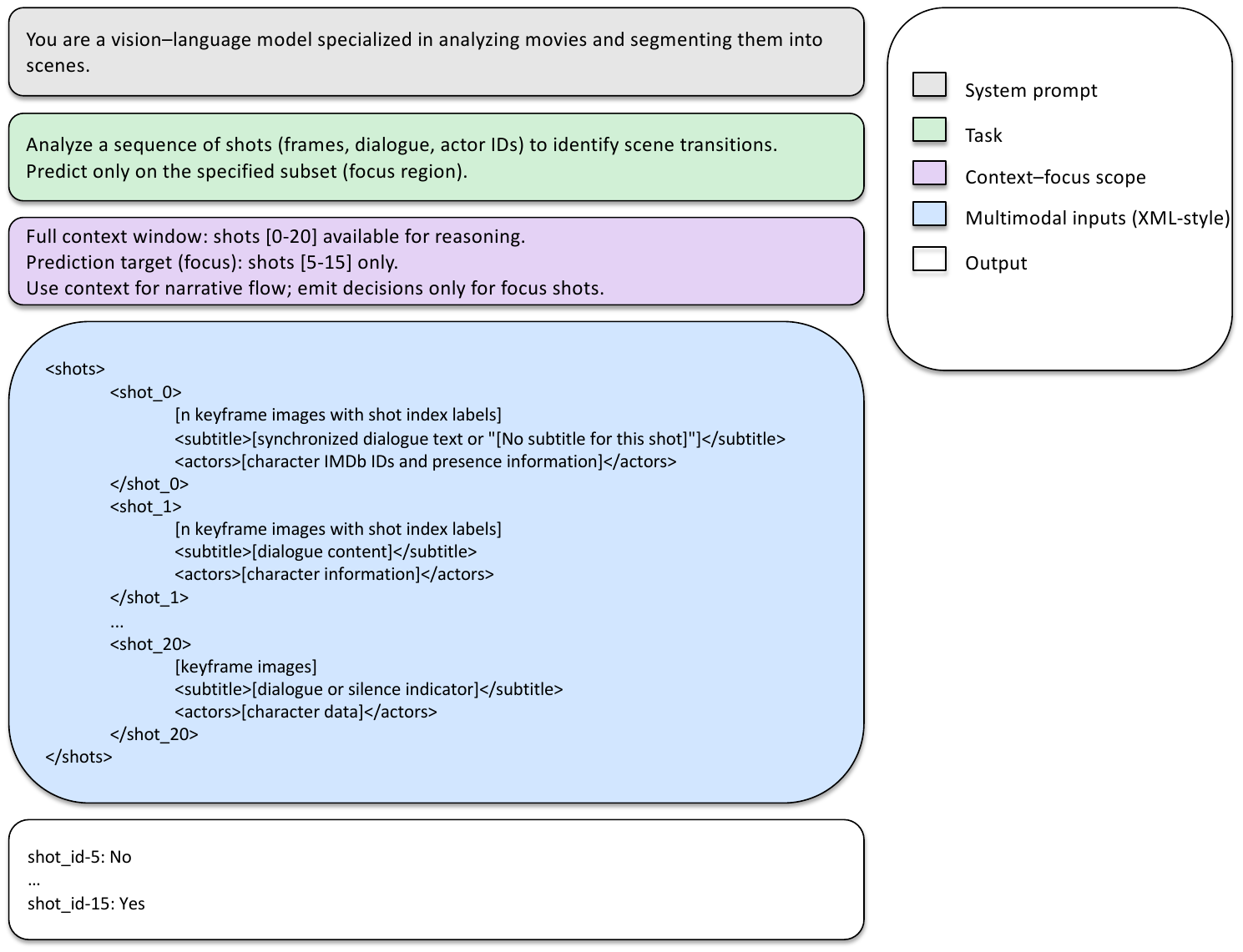}
    \caption{
    \textbf{Full prompt structure and output format.} Illustration of the multimodal prompt structure (instructions, visual frames, subtitles, and metadata) and the model's structured output with shot-level boundary predictions.
    }
    \label{fig:prompt_example}
\end{figure*}

\cref{fig:prompt_example} illustrates our full prompt structure and the expected output format. The \textcolor{black!60}{\textbf{gray}} block contains the \emph{system prompt}, the \textcolor{green!50!black}{\textbf{green}} block provides the \emph{task instructions}, the \textcolor{blue!60!black}{\textbf{blue}} block encodes the \emph{per-shot multimodal inputs} (frames, subtitles, actor IDs) in an XML-style layout, and the \textcolor{purple!60!black}{\textbf{purple}} block defines the \emph{context-focus scope} (indicating which shots in the context window are also prediction targets). As depicted in the output block, the model generates \verb|shot_id: Yes/No| decisions only for shots in the focus window. On average, a complete prompt contains $7939$ words.

\subsubsection{Visual Shot-ID Markers}
\label{sup:shot_id_details}
\begin{figure}[htpb]
    \centering
    \includegraphics[width=\linewidth]{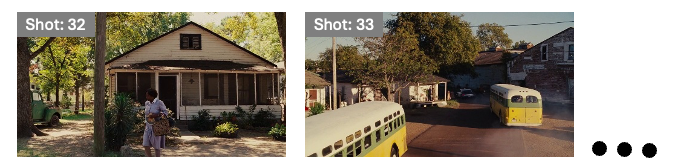}
    \caption{\textbf{Visual shot-ID markers.} A compact numerical tag is overlaid at the top-left corner of each frame, tightly coupling visual content with the corresponding textual shot IDs referenced in the prompt while preserving semantically relevant content.}
    \label{fig:visual_id_example}
\end{figure}

To strengthen the correspondence between visual frames and their textual shot identifiers in the prompt, we overlay a small, high-contrast numerical tag at the top-left corner of each frame, as illustrated in~\cref{fig:visual_id_example}. This visual marker serves two purposes: (1) it explicitly anchors each frame to its corresponding shot ID in the structured text input, reducing ambiguity during multimodal reasoning, and (2) it provides a consistent spatial reference that helps the model track shot boundaries across the temporal sequence. The marker is deliberately positioned to minimize occlusion of semantically relevant content while maintaining high visibility. As demonstrated in our ablation study (\cref{sec:input_ablt}), these markers improve the final results.

\subsection{Setup}
\subsubsection{Datasets}

\paragraph{Explainability.} To enable aligning our model to generate post-hoc rationales for its predicted scene boundaries (\cref{sec:explainability_res}), we curate a small supervision set of 35 samples pairing annotated scene boundaries with human-written explanations. Each entry contains the \texttt{movie\_id}, the inclusive \texttt{start\_shot} and \texttt{end\_shot} indices of the scene being concluded, and a free-text \texttt{rationale} describing the narrative transition. For example: \textit{“There is a clear scene transition: the narrative shifts from an interview between two women about one woman’s past to a sequence showing her working as a maid and caring for a child, changing place, time, and situation.”}

\subsubsection{Metrics}

\paragraph{Scene segmentation.}
Following prior work~\cite{sadoughi2023mega, islam2023efficient}, we report the following standard detection metrics on the MovieNet \cite{huang2020movienet} and BBC \cite{baraldi2015deep} datasets:
\begin{itemize}
    \item \textbf{Average Precision (AP)}: area under the precision-recall curve.
    \item \textbf{F1}: harmonic mean of precision and recall at an optimized operating point.
\end{itemize}

\paragraph{Video Chaptering.}
Following~\cite{ventura2025chapter,yang2023vidchapters}, we evaluate chapter segmentation and titling using:

\begin{itemize}
  \item \textbf{Chapter F1}: boundary-detection F1 computed against creator-provided chapter endpoints. Since endpoints are continuous timestamps, matching is based on temporal overlap and averaged over multiple overlap lengths; see~\cite{ventura2025chapter} for details.
  \item \textbf{tIoU}: temporal Intersection-over-Union between predicted and ground-truth chapters, measuring temporal alignment; the exact protocol follows~\cite{ventura2025chapter}.
  \item \textbf{SODA and CIDEr}: metrics for evaluating semantic alignment of generated titles to ground-truth titles (see \cite{yang2023vidchapters} for details).
\end{itemize}

\subsubsection{Additional Implementation Details}

\paragraph{Frame Processing.}
Our vision-language model \cite{qin2024qwen2_5vl} can accept a user defined image size. All frames across datasets are resized to \(147\times 63\) pixels. This resolution was selected to balance computational requirements with the preservation of visual details required for training and inference.

\paragraph{Training.}
All experiments were conducted on a cluster of 8$\times$A100 (40\,GB) GPUs. We fine-tune Qwen2.5\mbox{-}VL\mbox{-}7B on MovieNet ($\sim$29k samples) using LoRA \cite{hu2022lora} (rank~8, $\alpha$=16) for 4 epochs. With mixed precision, FlashAttention \cite{dao2022flashattention}, and ZeRO-3 \cite{rajbhandari2020zero} sharding (data/optimizer/parameter partitioning), end-to-end fine-tuning on MovieNet-318 completes in approximately 2–4 hours, and on the chaptering task in around 1 hour, depending on I/O and kernel availability.

\paragraph{Inference.}
Evaluation on the MovieNet test split takes approximately 1--2 hours on 8$\times$A100 (40GB) GPUs with data parallelism. Each movie is partitioned into non-overlapping context windows. We run batch-wise sequential decoding per window and then aggregate the outputs to compute the final metrics. This inference procedure applies to both tasks (scene segmentation and video chaptering). 

\subsubsection{Chapter-LLaMA Adaptation for MovieNet}
Adapting Chapter-LLaMA \cite{ventura2025chapter} to MovieNet is conceptually straightforward but requires addressing a modality gap. While Chapter-LLaMA relies on rich transcripts and keyframe captions, MovieNet, being cinematic, often exhibits sparser dialogue while scene changes are predominantly \emph{visual}. Hence, directly applying Chapter-LLaMA to MovieNet under-detects boundaries. For a fair comparison, we generate a caption \emph{per-shot} using the authors’ VLM \cite{yao2024minicpm} to supply the missing visual semantics, and feed these captions to Chapter-LLaMA in the \emph{exact} input format specified in~\cite{ventura2025chapter}. This preserves their pipeline while aligning the visual signal with MovieNet’s shot structure.

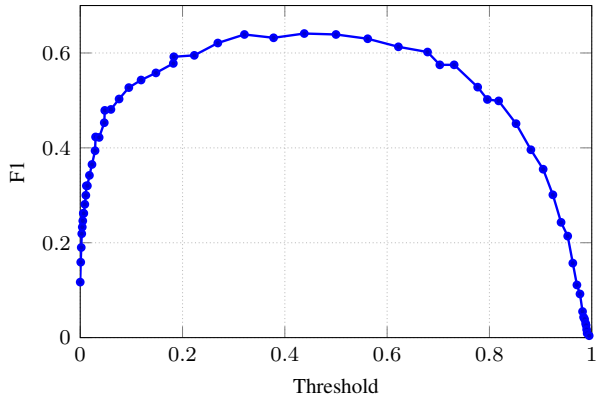
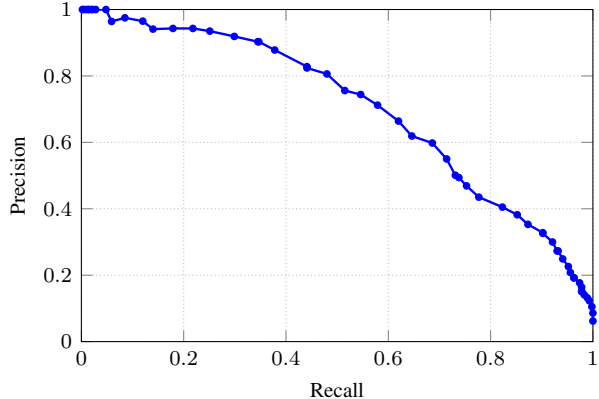
\begin{figure*}[t!]
\centering
\begin{subfigure}{0.48\textwidth}
\centering
\begin{tikzpicture}
\begin{axis}[
  width=\linewidth, height=6cm,
  xlabel={Threshold}, ylabel={F1},
  xmin=0, xmax=1, ymin=0, ymax=0.7,
  grid=both, grid style={densely dotted},
  tick label style={font=\footnotesize},
  label style={font=\footnotesize},
]
\addplot+[mark=*, mark size=1.2pt, line width=0.9pt] coordinates {
(0.000,0.117) (0.001,0.159) (0.002,0.190) (0.003,0.219) (0.004,0.233)
(0.005,0.246) (0.006,0.262) (0.007,0.262) (0.009,0.281) (0.011,0.300)
(0.012,0.320) (0.014,0.320) (0.018,0.342) (0.023,0.365) (0.029,0.394)
(0.030,0.423) (0.037,0.422) (0.047,0.453) (0.048,0.479) (0.060,0.481)
(0.076,0.503) (0.095,0.527) (0.119,0.543) (0.148,0.558) (0.182,0.578)
(0.183,0.592) (0.223,0.595) (0.269,0.621) (0.321,0.639) (0.378,0.632)
(0.438,0.641) (0.500,0.639) (0.562,0.630) (0.622,0.613) (0.679,0.602)
(0.703,0.575) (0.731,0.575) (0.777,0.528) (0.796,0.502) (0.818,0.499)
(0.852,0.451) (0.881,0.396) (0.905,0.355) (0.924,0.301) (0.940,0.243)
(0.953,0.214) (0.963,0.157) (0.971,0.111) (0.977,0.092) (0.982,0.055)
(0.984,0.043) (0.986,0.039) (0.988,0.030) (0.989,0.026) (0.990,0.017)
(0.991,0.009) (0.995,0.004)
};
\end{axis}
\end{tikzpicture}
\caption{
\textbf{F1 versus decision threshold.} Peak \(\mathrm{F1}=0.641\) at threshold \(\approx0.438\). Broad plateau \(\approx0.62\)--0.64 indicates stable operating range.
}
\label{fig:f1_vs_threshold_side}
\end{subfigure}
\hfill
\begin{subfigure}{0.48\textwidth}
\centering
\begin{tikzpicture}
\begin{axis}[
  width=\linewidth, height=6cm,
  xlabel={Recall}, ylabel={Precision},
  xmin=0, xmax=1, ymin=0, ymax=1,
  grid=both, grid style={densely dotted},
  tick label style={font=\footnotesize},
  label style={font=\footnotesize},
  legend style={draw=none, font=\footnotesize, at={(0.03,0.97)}, anchor=north west}
]
\addplot+[mark=*, mark size=1.0pt, line width=0.9pt] coordinates {
(1.000,0.062) (1.000,0.086) (0.998,0.105) (0.993,0.123) (0.989,0.132)
(0.983,0.141) (0.978,0.151) (0.978,0.151) (0.978,0.164) (0.974,0.177)
(0.963,0.192) (0.963,0.192) (0.956,0.208) (0.952,0.226) (0.941,0.249)
(0.932,0.273) (0.930,0.273) (0.921,0.300) (0.902,0.326) (0.902,0.328)
(0.873,0.353) (0.852,0.382) (0.823,0.405) (0.777,0.435) (0.753,0.469)
(0.738,0.494) (0.731,0.501) (0.714,0.550) (0.686,0.598) (0.646,0.619)
(0.620,0.664) (0.579,0.712) (0.546,0.744) (0.515,0.756) (0.480,0.806)
(0.441,0.824) (0.441,0.828) (0.378,0.878) (0.347,0.903) (0.345,0.903)
(0.299,0.919) (0.251,0.935) (0.218,0.943) (0.179,0.943) (0.140,0.941)
(0.120,0.965) (0.085,0.975) (0.059,0.964) (0.048,1.000) (0.028,1.000)
(0.022,1.000) (0.020,1.000) (0.015,1.000) (0.013,1.000) (0.009,1.000)
(0.004,1.000) (0.002,1.000)
};
\end{axis}
\end{tikzpicture}
\caption{
\textbf{Precision--Recall curve.} Showing a balanced knee. E.g., at \(t\approx0.321\): \(P\approx0.60\), \(R\approx0.69\); at \(t\approx0.50\): \(P\approx0.71\), \(R\approx0.58\).
}
\label{fig:pr_curve_side}
\end{subfigure}

\caption{
\textbf{F1 and Precision--Recall curves across decision thresholds.} Both curves demonstrate strong operating flexibility for task-dependent tuning.
}
\label{fig:threshold_curves_side_by_side}
\end{figure*}

\section{Additional Experiments}
\subsection{Alternative Prediction Schemes}
\label{sup:confidence_score}

Confidence scores are essential for scene segmentation, enabling flexible precision–recall trade-offs across different operating points. However, extracting reliable confidence from VLM outputs is non-trivial, as VLMs produce structured textual responses where confidence must be inferred from token-level probabilities rather than dedicated classification heads. In this section, we compare our proposed approach against two alternative prediction schemes, which differ in output format, confidence estimation capability, accuracy, and computational cost.

\vspace{-3mm}
\paragraph{Comprehensive scheme.}
Our proposed scheme (see ~\cref{sec:confidence_method}) generates structured outputs in the format \texttt{shot\_id:<id>: Yes/No} for each shot in the focus window, providing \emph{explicit} predictions for all target shots. Confidence is then computed from the normalized token logits as described in~\cref{eq:conf}. As shown in~\cref{tab:output_formats}, this scheme achieves strong performance (F1: 62.1, AP: 66.8), but requires an average inference time of 87.2 seconds per movie.

\vspace{-4mm}
\paragraph{Concise scheme.}
While our Comprehensive scheme is accurate, its inference latency may be prohibitive for latency-sensitive applications. To address this, we explore a more efficient output format where the model emits only \texttt{Yes} tokens for detected boundaries, omitting explicit \texttt{No} predictions for non-boundary shots. However, a subtle question then arises: can we extract confidence scores from this format by probing the \texttt{Yes} token logits? Unfortunately, this approach is fundamentally flawed due to the model's output structure. When the model predicts a shot ID, it has already committed to marking that shot as a boundary, which means that the subsequent \texttt{Yes} token is obligatory rather than a genuine binary choice. Consequently, $p(\texttt{Yes} \mid \text{shot\_id\_predicted}) \approx 1$ by design, making the confidence score uninformative. This contrasts sharply with the Comprehensive scheme, where the \texttt{Yes}/\texttt{No} choice represents a genuine binary decision with meaningful probability mass on both outcomes.

Given this limitation, we evaluate the Concise scheme \textbf{without confidence extraction}, using the model's outputs directly: a shot is classified as a boundary if and only if the model predicts \texttt{Yes} for it. Despite lacking precision–recall flexibility, this variant achieves competitive performance (F1: 53.4) with recent methods such as TranS4mer (48.4) and MEGA (55.3), while delivering dramatic speedup over the Comprehensive scheme (10.5s per movie, approximately 8.3$\times$ faster). This makes it an attractive option when inference speed is critical and precision–recall control is not required.

\vspace{-3mm}
\paragraph{Concise scheme with repeated sampling.}
Given the inability to extract meaningful confidence using the Concise scheme, we investigate whether repeated sampling can provide reliable confidence estimates. Specifically, we perform $m{=}5$ independent inference runs, draw temperatures uniformly from the interval $[0.5,1.0]$ for each run, and compute confidence per shot as the proportion of \texttt{Yes} outcomes. Unfortunately, as shown in~\cref{tab:output_formats}, this approach fails on both fronts: it is less accurate than Concise without confidence (F1: 52.6 vs. 53.4) and even slower than the Comprehensive scheme (105.2s vs. 87.2s), making it strictly worse than both alternatives.

To summarize, the \textbf{Comprehensive scheme} is recommended when accuracy and precision–recall control are paramount, while the \textbf{Concise scheme (without confidence)} may be a good alternative when inference speed is the priority and precision–recall control is not strictly required.

\begin{table}[htpb]
  \caption{\textbf{Comparison of prediction schemes for scene segmentation.} The Comprehensive scheme achieves the best accuracy and features a confidence prediction capability, while the Concise scheme (without confidence) offers a compelling speed-accuracy trade-off for latency-critical applications.}
  \vspace{-2mm}
  \centering
  \footnotesize
  \resizebox{\columnwidth}{!}{%
  \begin{tabular}{l|ccc}
    \toprule
    \textbf{Prediction Scheme} & \textbf{F1}~$\uparrow$ & \textbf{AP}~$\uparrow$ & \textbf{Avg.\ time / movie (s)}~$\downarrow$ \\
    \midrule
    Concise (without confidence)            & 53.4 & - & \textbf{10.5} \\
    Concise (repeated sampling) & 52.6 & 34.7 & 105.2 \\
    Comprehensive            & \textbf{62.1} & \textbf{66.8} & \underline{87.2} \\
    \bottomrule
  \end{tabular}}
  \vspace{-1mm}
  \label{tab:output_formats}
\end{table}

\subsection{Model F1 and Precision--Recall Analysis}
To assess our method's sensitivity to threshold changes, we plot F1 versus decision threshold (\cref{fig:f1_vs_threshold_side}) alongside the corresponding precision--recall curve (\cref{fig:pr_curve_side}). As depicted, the F1 curve rises sharply from near-zero thresholds and reaches a broad plateau, peaking at F1 $=0.641$ around threshold 0.438. This plateau (\(\approx\)0.62--0.64 F1 across thresholds \(\sim\)0.27--0.56) demonstrates stable performance with minimal sensitivity to threshold variations. The PR curve exhibits the expected trade-off between precision and recall. For \emph{balanced} operation, a threshold near 0.321 yields \(\mathrm{P}\approx 0.60\) and \(\mathrm{R}\approx 0.69\) (\(\mathrm{F1}=0.639\)). For \emph{precision-oriented} applications, a threshold near 0.50 yields \(\mathrm{P}\approx 0.71\) and \(\mathrm{R}\approx 0.58\) (\(\mathrm{F1}=0.639\)). Conversely, recall-oriented scenarios can use lower thresholds (e.g., 0.223--0.269) to push recall above 0.70 with modest precision. To summarize, these curves demonstrate that flexible task-dependent tuning is possible while maintaining robust performance across the F1 plateau.

\subsection{Zero-Shot VLM Baselines}
\label{sup:zero_shot_baselines}
To assess the necessity of fine-tuning, we evaluate zero-shot performance on MovieNet using two VLMs: Qwen2.5-VL-7B~\cite{bai2025qwen2} (our base model without fine-tuning) and Claude 4.5 Sonnet~\cite{claude_sonnet_4_5} (a strong closed-source VLM). Due to Claude API constraints, we use 1 frame per shot for all methods in this comparison. Since zero-shot models do not reliably follow the required output format, we use LLM-based parsing to extract their predictions and treat unparsed shots as negatives. As shown in~\cref{tab:zero_shot_baselines}, zero-shot methods significantly underperform our fine-tuned model. Qwen2.5-VL-7B achieves only 11.1 F1 with a 7.9\% parse error rate, while Claude 4.5 Sonnet reaches 37.6 F1 with near-zero parse errors (0.03\%). In contrast, Scene-VLM achieves 61.8 F1 with zero parse errors using simple rule-based parsing. Moreover, zero-shot methods cannot provide meaningful confidence scores for their predictions: for closed-source models (e.g., Claude) logits are inaccessible; for Qwen, LLM-based parsing makes confidence extraction infeasible while rule-based parsing fails 71\% of the time (not shown). These results confirm that fine-tuning is essential for reliable scene segmentation.

\begin{table}[htpb]
  \caption{\textbf{Zero-shot VLM baselines on MovieNet.} Fine-tuning is essential: zero-shot methods significantly underperform and cannot provide confidence scores. All methods use 1 frame per shot.}
  \centering
  \resizebox{\columnwidth}{!}{%
  \begin{tabular}{@{}lccccc@{}}
    \toprule
    \textbf{Method} & \textbf{Fine-Tuned} & \textbf{Parse} & \textbf{Parse Err.}~$\downarrow$ & \textbf{F1}~$\uparrow$ & \textbf{AP}~$\uparrow$ \\
    \midrule
    Qwen2.5-VL-7B~\cite{bai2025qwen2} & $\times$ & LLM & 7.9\% & 11.1 & -- \\
    Claude 4.5 Sonnet~\cite{claude_sonnet_4_5} & $\times$ & LLM & 0.03\% & 37.6 & -- \\
    \midrule
    \textbf{Scene-VLM (7B)-1F (ours)} & $\checkmark$ & Rule & \textbf{0\%} & \textbf{61.8} & \textbf{65.3} \\
    \bottomrule
  \end{tabular}}
  \label{tab:zero_shot_baselines}
\end{table}

\begin{table}[t]
  \caption{\textbf{Computational analysis.} Peak memory, latency, and accuracy at 10 samples. “F” denotes frames per shot. Mean and standard deviation values are computed over five runs.}
  \centering
  \normalsize
  \resizebox{\columnwidth}{!}{%
  \begin{tabular}{lccccc}
    \toprule
    Method & Params & Memory (GB) $\downarrow$ & 10-sample latency (s) $\downarrow$ & F1 $\uparrow$ & AP $\uparrow$ \\
    \midrule
    TranS4mer-3F~\cite{islam2023efficient} & 37M & \textbf{1} & \textbf{0.24 $\pm$ 0.0} & 48.4 & 60.8 \\
    \midrule
    Scene-VLM-1F   & 3B & 7  & $1.15 \pm 0.07$ & 55.7 & 58.2 \\
    Scene-VLM-3F   & 3B & 9  & $1.35 \pm 0.08$ & 59.6 & 62.8 \\
    Scene-VLM-1F   & 7B & 16 & $1.84 \pm 0.15$ & 61.8 & 65.3 \\
    Scene-VLM-3F   & 7B & 18 & $2.34 \pm 0.14$ & \textbf{62.1} & \textbf{66.8} \\
    \bottomrule
  \end{tabular}}
  \label{tab:comp-analysis}
  \vspace{-4mm}
\end{table}

\subsection{Computational Analysis}
\label{sup:compute_analysis}
In this section, we present a thorough analysis of our models' computational requirements. We compare four variants, varying the number of frames per shot (1F or 3F) and model size (3B or 7B). For all models, we use a batch size of 1 to enable a fair comparison of memory and runtime. We do not apply model/system optimizations such as weight/activation quantization (e.g., 4-/8-bit) or inference engines with KV-cache optimizations (e.g., \texttt{vLLM}); these could further reduce both latency and memory in future work.

We report peak memory, latency, and accuracy at \emph{10 samples} (i.e., 10 binary boundary decisions) in \cref{tab:comp-analysis}, using the same configuration as in the main experiments: a context window of 20 shots and a focus window of 10 shots. We repeat the evaluation five times, reporting the mean and standard deviation for wall-clock latency, and the maximum over runs for peak memory. As depicted, reducing frames per shot from 3F to 1F in the 7B model lowers latency from 2.34\,s to 1.84\,s (\(\approx\)21\%) and peak memory from 18\,GB to 16\,GB, with only a minor accuracy drop (F1/AP: 62.1/66.8 \(\rightarrow\) 61.8/65.3); a similar trend holds for the 3B model. The latency reduction stems from the fact that frames dominate the token count of the input, so removing two of three frames per shot shortens the length of the multimodal input sequence and reduces computation. Meanwhile, the small accuracy drop aligns with our results from \cref{sec:num-frames-per-shot}, which shows that performance degrades modestly when reducing the number of frames per shot. Intuitively, since shots are segments which typically contain no major visual changes, a single representative frame often preserves most scene-transition-relevant information.

We also compare against TranS4mer~\cite{islam2023efficient} (MEGA~\cite{sadoughi2023mega} has not released source code), using the same 3 frames per shot configuration for a fair comparison. As shown in~\cref{tab:comp-analysis}, TranS4mer is faster and more memory-efficient (37M parameters, 1\,GB memory, 0.24\,s latency), but Scene-VLM offers significantly higher accuracy (+13.7 F1) along with explainability capabilities that are unavailable in encoder-based methods. Notably, our smaller 3B variant narrows this efficiency gap while still outperforming TranS4mer by +11.2 F1. Finally, we note again that no inference optimizations were applied to our models; such techniques could further reduce our latency and memory.

\subsection{Explainability for Scene Segmentation (Cont.)}
\label{sec:cont_explainability}
\begin{figure*}[t]
    \centering
    \includegraphics[width=1\linewidth]{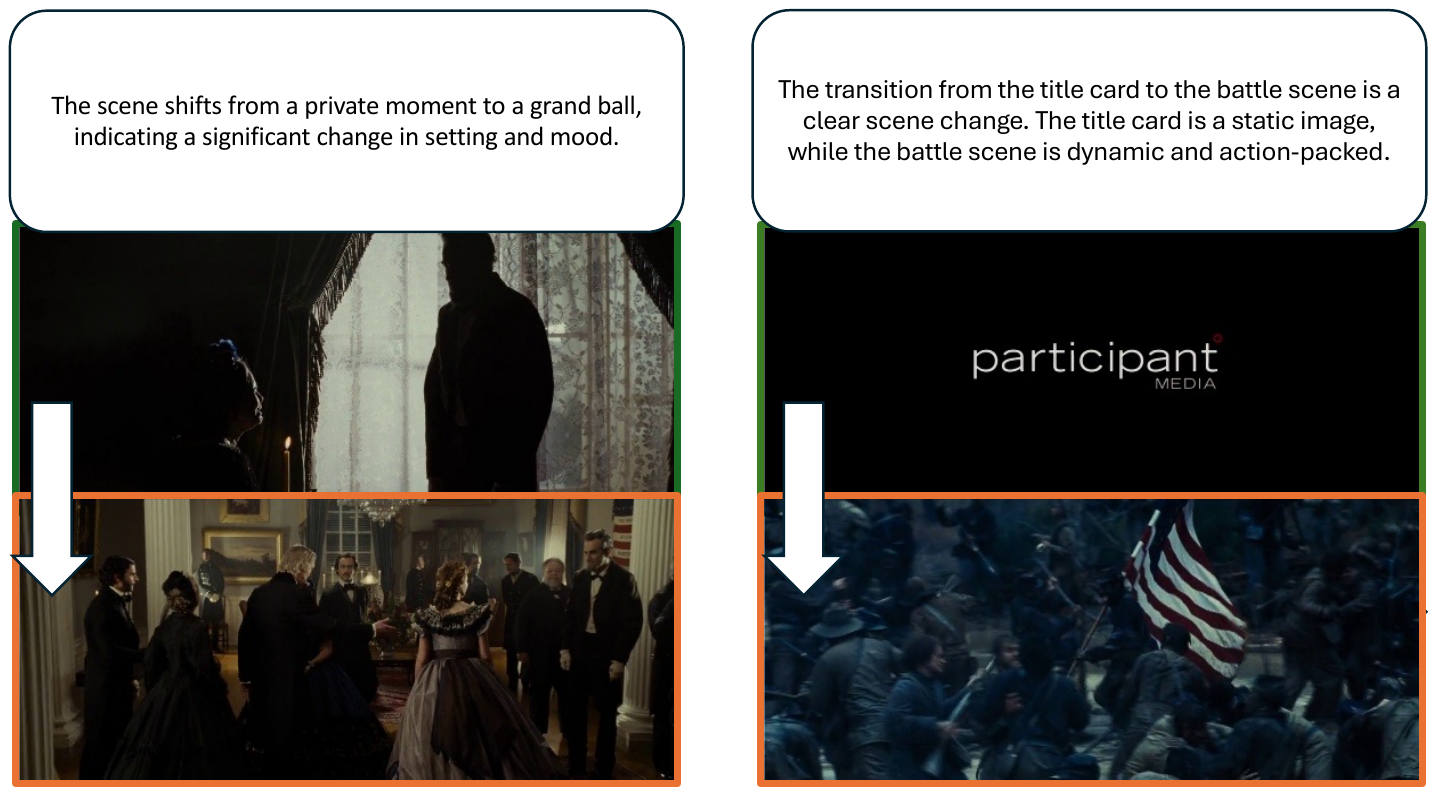}
    \vspace{-30mm}
    \caption{\textbf{Example 1.} \emph{Left:} boundary due to a location change. \emph{Right:} transition from a title card to the opening shot.}
    \label{fig:app-explain-2}
\end{figure*}

\begin{figure*}[t]
    \centering
    \includegraphics[width=1\linewidth]{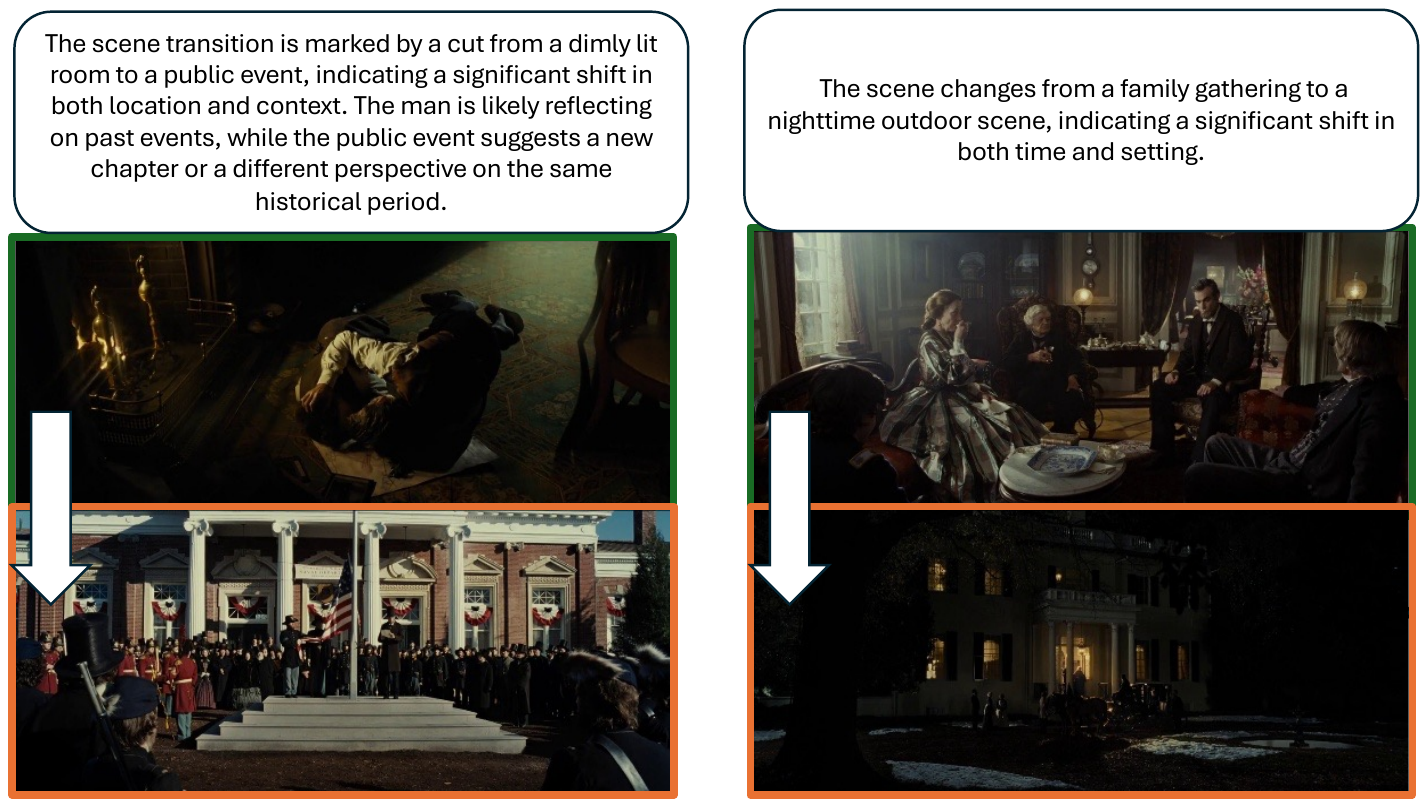}
    \vspace{-30mm}
    \caption{\textbf{Example 2.} \emph{Left and right panels:} boundary justified by a joint change in \emph{time} and \emph{place}; the model references visual cues (lighting, background) and/or dialogue context to support the decision.}
    \label{fig:app-explain-1}
\end{figure*}

We present additional qualitative examples of model-generated rationales for scene-boundary decisions in~\cref{fig:app-explain-2} and~\cref{fig:app-explain-1}. These examples span both abrupt visual transitions (e.g., title cards) and subtler, socially driven changes (e.g., shifts in location, time, or conversational structure). 

\vspace{-300mm}

\end{document}